\documentclass[preprint,3p,times,a4paper,authoryear]{elsarticle}

\usepackage{graphicx}
\usepackage{amssymb}
\usepackage{lineno}

\usepackage{url}
\usepackage{makecell}
\usepackage{bm}
\usepackage{mathtools}

\usepackage{caption}
\usepackage{subcaption}
\usepackage{tabularx}

\usepackage{xcolor}
\usepackage[normalem]{ulem}

\journal{Elsevier}

\begin{document}

\begin{frontmatter}


\title{Introducing Super Pseudo Panels: Application to Transport Preference Dynamics}


%\author[label1,label2]{<author name>}
%\address[label1]{<address>}
%\address[label2]{<address>}

\author{Stanislav S. Borysov}
\ead{stabo@dtu.dk}
\author{Jeppe Rich}
\address{Department of Management, Technical University of Denmark, DTU, 2800 Kgs. Lyngby, Denmark}

\begin{abstract}
We propose a new approach for constructing synthetic pseudo-panel data from cross-sectional data. The pseudo panel and the preferences it intends to describe is constructed at the individual level and is not affected by aggregation bias across cohorts. This is accomplished by creating a high-dimensional probabilistic model representation of the entire data set, which allows sampling from the probabilistic model in such a way that all of the intrinsic correlation properties of the original data are preserved. The key to this is the use of novel deep learning algorithms based on the Conditional Variational Autoencoder (CVAE) framework. From a modelling perspective, the concept of a model-based resampling creates a number of opportunities in that data can be organized and constructed to serve very specific needs of which the forming of heterogeneous pseudo panels represents one. The advantage, in that respect, is the ability to trade a serious aggregation bias (when aggregating into cohorts) for an unsystematic noise disturbance. Moreover, the approach makes it possible to explore high-dimensional sparse preference distributions and their linkage to individual specific characteristics, which is not possible if applying traditional pseudo-panel methods.  

We use the presented approach to reveal the dynamics of transport preferences for a fixed pseudo panel of individuals based on a large Danish cross-sectional data set covering the period from 2006 to 2016. As the resampling of the preferences is conditional on the specific socio-economic profiles, sampling noise that normally arises from collecting repeated cross-sectional data can be eliminated. The model is also utilized to classify individuals into 'slow' and 'fast' movers with respect to the speed at which their preferences change over time. It is found that the prototypical fast mover is a young woman who lives as a single in a large city whereas the typical slow mover is a middle-aged man with high income from a nuclear family who lives in a detached house outside a city.
\end{abstract}


\end{frontmatter}

%\linenumbers


%-------------------------------------------------------------------------
\section{Introduction}
\label{sec:intro}
%-------------------------------------------------------------------------

Understanding preferences and their dynamics, whether these are related to transport or any other domain, is a fundamental research question which have impact not only for models and predictions but also for the way policies are designed and towards whom they should be targeted. It is beyond doubt that habituates and preferences change over time \citep{Garling2003} and affect transport behaviour in significant ways. Examples include \cite{VIJ2017238} who consider modal preference shifts in the San Francisco area, the understanding of how value-of-time preferences change through a financial crisis as considered in \cite{richvandet2019}, understanding of technology uptake \citep{MAU2008504} and the dynamics of car ownership \citep{Cirillo2015, NOLAN2010446} to mention a few. 

Two model approaches remain popular for estimating dynamic behaviour: (i) panel methods \citep{KITAMURA1990401} and (ii) pseudo-panel methods \citep{DEATON1985109}. Whereas the native panel approach has a number of theoretical advantages over pseudo-panel methods \citep{hsiao_2014}, it is often faced with severe practical challenges related to the collection of data. Particular challenges include the occurrence of selective attrition \citep{KAPLAN1987297} and response fatigue \citep{Golob1997}. The former refers to the case where participants drop out of a study in a predictable way (e.g., low socio-economic status participants may be more likely to drop out compared to other participants). The latter refers to a general problem of collecting data repeatedly over time as the drop out rate becomes a function of the panel length. This generally causes longitude panels to be both expensive and limited in the number of recurring respondents they cover, and this has led to an increasing interest in pseudo-panel methods. 

Pseudo-panel models are typically based on repeated cross-sectional data \citep{wooldridge2010econometric}. Whereas pseudo-panel methods overcome many of the challenges related to the collecting of panel data, they imply other limitations for the model framework. Where panel data can take explicit account of individual-specific heterogeneity by operating at the level of individuals, pseudo panels involve aggregation and thereby invoke the assumption of 'representative individuals'. This aggregation into prototypes of individuals is generally referred to as 'cohorts'. Typically, these cohorts are identified by characteristics that remain stable over time (such as the gender, geographical region or year of birth). In the estimations, therefore, it becomes possible to capture fixed 'cohort effect' which would otherwise result in biased estimations. As a result, the methodology is essentially an instrumental variables approach in which cohort dummies are used as instruments. Pseudo panels have been used extensively in the economic literature to examine a variety of subjects from consumption dynamics \citep{Gardes2005}, long-term behavioural changes related to wage dynamics \citep{DELLAS2003843}, labour market participation \citep{Afsa2006} and earning mobility \citep{Antman2005}. In the transport community, they have been used mainly to explore car ownership dynamics as presented in \cite{dargay1999}, \cite{DARGAY2002351} and later in \citet{Huang2007}. 

When constructing cohorts, the modeller faces a classical trade-off between the number of cohorts and the number of individuals within each cohort \citep{Gardes2005}. If individuals are allocated into a large number of cohorts, there will be few observations in each cohort which generally increase the heterogeneity between cohorts at different times. This is referred to as 'measurement errors' in \cite{DEATON1985109} and generally leads to biased parameter estimates. A way to resolve this problem is to use large samples within cohorts as proposed by \cite{DEATON1985109}. However, as argued in \cite{VERBEEK1993125}, often quite a substantial number of observations are required. This, in turn, creates another problem that results in loss of efficiency of the estimators \citep{Cramer1964}. Thus, the challenge when constructing a pseudo panel is finding a balance between the number of cohorts and the number of individuals within cohorts. There is rich literature that considers the problem of heterogeneity and its impact on model properties \citep{Granger1988,im2003testing}, and also how predictions and derived policies could be misleading \citep{Hsiao2005}. If individuals are truly heterogeneous, not only can the time series properties of aggregate data be substantially different from those of disaggregate data (e.g. \cite{Granger1988,Lewbel1994,im2003testing}), the evaluation of policies based on aggregate data may be equally misleading. Another concern is that predictions of aggregate outcomes using aggregate data will be less accurate than predictions based on a bottom-up approach (e.g. \cite{Hsiao2005}). 

From the perspective of investigating preferences over time, pseudo panels (despite all of the aggregation and heterogeneity challenges described above) have one major advantage compared to panels. A challenge with panels is that, although individuals and households can be tracked accurately over time, life events such as marriage, kids, residential relocation and change of car ownership status will affect the preference distribution from one time period to the next. This has been evidenced in a number of papers including \cite{DEHAAS2018140}, \cite{SCHOENDUWE201598} and \cite{MUGGENBURG2015151}. It is a general conclusion that, for all life events, significant effects on the mobility pattern are found. On the one hand, this is interesting as it suggests that there might indeed be windows of opportunity to change travel behaviour when a life event occurs \citep{DEHAAS2018140}. However, it also introduces the problem of disentangling possible causes for a given mobility pattern. In other words, too much is going on and it is non-trivial to filter out the different parts. For instance, when considering preference changes, it is generally difficult to compare preferences for individuals in completely different settings as the preference change may have been derived from these changed settings. A more 'clean' comparison is to adopt the pseudo-panel way of thinking and move the same set of individuals framed in exactly the same socio-economic setting forward in time while simultaneously exploring their preference distribution. 

\subsection{The approach}
\label{sec:intro:approach}

\begin{figure}[t]
\centering
\includegraphics[height=5.0cm]{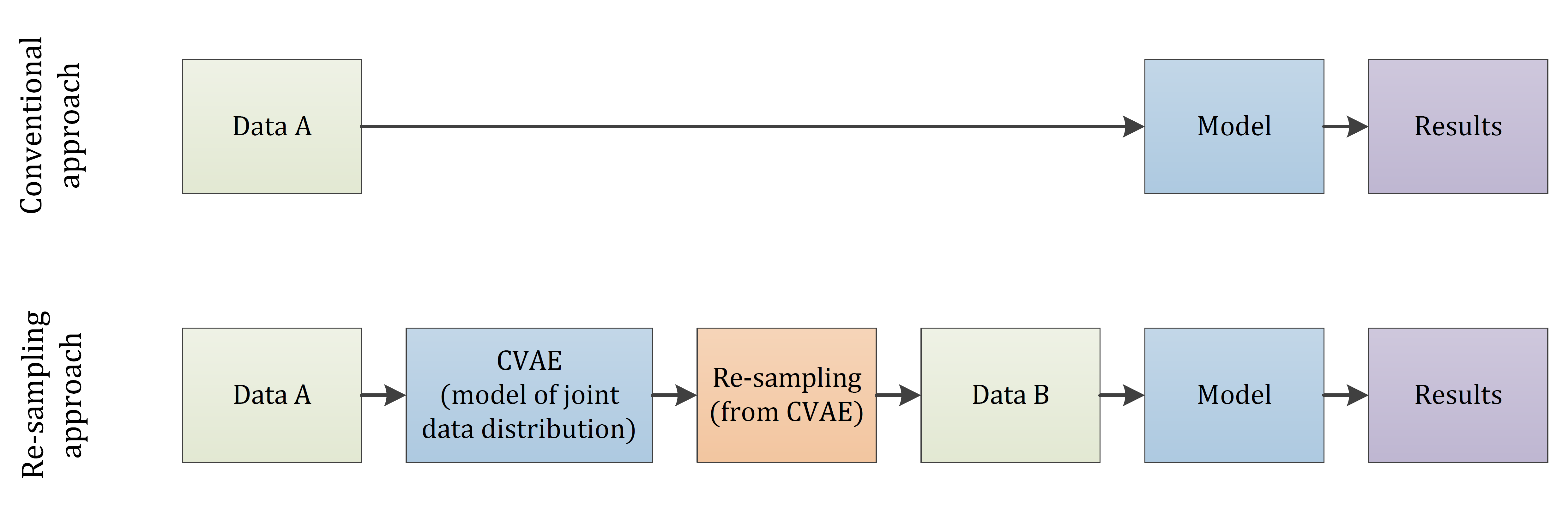}
\caption{Illustration of the CVAE resampling approach.}
\label{fig:pp_resample}
\end{figure}

The aim of the paper is to facilitate the aforementioned analysis by introducing the concept of 'super pseudo panels' (SPPs), which is an attempt to circumvent the problem of aggregation bias in conventional pseudo-panels. Or, more precisely, trade the better known aggregation bias for a different type of bias which is expected to be less severe and under the control of the modeller. This is achieved by utilizing newly developed machine learning algorithms which can mimic the properties of high-dimensional data. The models adopt a deep generative modelling approach \citep{Goodfellow-et-al-2016} from machine learning based on the Variational Autoencoder (VAE) model \citep{kingma2013auto}. The benefit of this approach is that the model can act as a 'sampler' of individuals in such a manner that all of the intrinsic correlation properties of the original high-dimensional data are preserved. A high-level illustration of the resampling approach is illustrated in Fig.~\ref{fig:pp_resample}. The idea is that, rather than working with the original data, which involve a range of limitations, it is instead possible to work with ``modelled'' augmented versions of data.  

This brings about a number of new possibilities in other areas in addition to the application to pseudo panels, for instance generation of synthetic populations aimed at agent-based modelling \citep{2018arXiv180806910B} and  tackling of data privacy issues when publishing or using data \citep{pmlr-v68-choi17a}. The approach can be applied to reveal transport preference dynamics over time from pure cross-sectional data. This is accomplished by using an extension of the VAE framework---the Conditional VAE (CVAE) introduced in \cite{KingmaCVAE} and \cite{SohnCVAE}---in combination with a decomposition of a travel survey participants into three attribute spaces: (i) a socio-economic description of individuals, (ii) a transport preferences space, and (iii) an attribute space that represents changes in the infrastructure and the economy. The approach starts by approximating the distribution of preferences conditional on socio-economic variables, changes to the infrastructure and time of the cross-sectional data. Then, based on a reference pool of individuals in a starting year, it is possible to sample preferences for the exact same pool of individuals in subsequent years. In other words, it becomes possible to move a given pseudo panel of individuals forward in time and investigate how their transport preferences evolve.

\subsection{Contribution of the paper}
\label{sec:intro:contrib}
The paper contributes to the methodological domain of transport-related data analytics by introducing the concept of aggregation-free pseudo panels that allow a detailed investigation of preference dynamics. This is accomplished by constructing detailed probabilistic models of data which are able to mimic the joint distribution of the data generating process (DGP). The key to this is the use of the CVAE framework that enables the modeller to move a specific panel of individuals forward in time and explore their preference distribution conditional on their socio-economic profile. By applying the method to the Danish cross-sectional data from 2006 to 2016, a number of interesting empirical findings are revealed that supplement and extent previous investigations of transport preferences dynamics in \cite{ELZARWI2017207}, \cite{Xiong2015} and \cite{GOULIAS1999535}. First and foremost, we classify individuals as 'slow' and 'fast' movers over the considered period. Slow movers are defined as those individuals which change their preferences the least. Fast movers are those individuals who change their preferences the most. This is accomplished by applying a metric on the preference distribution in order to assess which preference distributions change the most from 2006 to 2017. This in turn leads to the following observations that are believed to be of interest to the transport community in general:
\begin{enumerate}
\item Indication that the prototypical fast movers are young single female adults living in cities, whereas slow movers are middle-aged men with high incomes that live in non-single households outside the cities and in privately owned detached houses. 
\item Suggestion that age may not be the most critical classifier of being a slow mover, as elderly people over the age of 70 are changing preferences at a faster pace than middle-aged people.
%\item Indication that longer education is \sout{positively} {\bf negatively?} correlated with the probability of being a fast mover.
\item Suggestion that being employed and having higher income are positively correlated with the probability of being a slow mover.
\item Confirmation that fast movers live predominantly in cities.
\end{enumerate}

Clearly, the possibilities of analysing preference changes are almost unlimited and could include a more detailed analysis of social and spatial clustering. For instance, it might be possible to identify certain groups of people for which it would be easier to influence behaviour. This, in turn, could have consequences as to whom certain policies should be targeted. An interesting and closely related application is to apply the approach to marketing data to identify the consumers that may be more in favour of accepting a new product. However, a full coverage of all possible applications is beyond the scope of the current paper. 

\subsection{How the paper is organized}
\label{sec:intro:org}
The paper proceeds with outlining the conceptual framework of the SPP in Section~\ref{sec:frame}, whereas in Section~\ref{sec:cvae}, the CVAE framework is presented. Section~\ref{sec:case} presents a case study where preference dynamics is investigated. Finally, we summarize and offer conclusions in Section~\ref{sec:conclusion}.

%-------------------------------------------------------------------------
\section{Conceptual framework of SPPs}
\label{sec:frame}

A simple way of understanding the conceptual framework of the SPP is to consider a definition of the data it applies to. Consider a repeated cross-sectional data with $N_t$ individuals defined by their socio-economic profiles $s_{t,i}$ ($i=\overline{1,N_t}$) and preferences $v_{t,i}$ for the time $t$ ($t=\overline{0,T}$) where $T+1$ represents the number of time periods for which the survey has been collected. Each $s_{t,i}$ can be represented as a collection of $M_s$ socio-economic attributes $s_{t,i,j}$ ($j=\overline{1,M_s}$) and $v_{t,i}$ as a collection of $M_v$ preference attributes $v_{t,i,j}$ ($j=\overline{1,M_v}$).

%The latter could include such elements as infrastructure, the cost of travelling and the technological state of a given point in time. To some extent, the revealed preferences can be considered as a remainder of elements that are not, or cannot be included in $x_{t,i}$. Hence, by including infrastructure, transport cost and rising income in $x_{t,i}$, preferences are measured net of these elements and could include, but not limited to, preference changes resulting from changes in technology and the way we adopt these.%

Most traditional parametric models operate under the assumption that the data on which they are estimated represent a 'ground truth'. Although this, in many ways, is a reasonably approach in that the data represent true observations, it can be argued that this way of thinking is naive because any collection of data is subjected to randomness in almost every aspects of the data collection process. Hence, randomness occurs in the sample selection phase, in responses and in the way questions are perceived. As a result, a more appropriate and realistic assumption is to assume that the real population is a realization from a ground truth population distribution. In other words, any transport diary for a given year is one of many possible realizations of transport diaries. Had the data been collected by another data collector with a different sample protocol, the result would have been different. This way of thinking supports the idea put forward in the present paper. Namely, the assumption that the available data is a realization from an underlying joint distribution, $s_{t,i}, v_{t,i} \sim P(S, V|x_{t,i}, t)$, where $S$ and $V$ are random variables of socio-economic profiles and transport-related preferences and $x_{t,i}$ is a measure of external information for each individual. By design, because the same pool of individuals is moved forward, preferences are not affected by life events that happen over time and sample noise resulting from the different cross-section samples. Clearly, from a stringent statistical perspective, the approach cannot add more information than is already embedded in the original sample. The advantages, however, are that we are able to present and align this information in different ways. Also, the method may not be able to answer all questions related to preferences as, in principle, the joint distribution is so large (and sparse) that it is not supported by data everywhere. However, by using latent variables in combination with neural networks, the model indirectly imputes missing observations and provides the most likely probability distribution for these. This in turn makes it possible to investigate and screen preferences in an unforeseen systematic way across many dimensions. The information gathered during this process may supplement and support other existing approaches.

Knowing the joint distribution $P(S, V|x_{t,i}, t)$ gives rise to a lot of opportunities when analyse (and synthesize) populations and their preferences over time. In this paper, we focus on the problem of generating pseudo-panel (longitudinal) data for a number of individuals with the socio-economic profiles $s_{0,i} \equiv s_{t=0,i}$ fixed at time $t=0$. In contrast to the common approach where a single instance of preferences is analysed conditional on the data, that is $v_{t,i} \sim P(V|S=s_{0,i}, x_{t,i}, t)$, the probabilistic framework allows a more general approach where it is possible to analyse the entire distribution of preferences for each individual, $P_{t,i}(V) \equiv P(V|S=s_{0,i}, x_{t,i}, t)$. Clearly, this assumes that the conditional distribution $P(V|S, x_{t}, t)$ can be estimated from the data in a sufficiently effective manner. When both $M_s$ and $M_v$ are small, several model approaches from generative modeling  are available, for example, based on traditional probabilistic graphical models \citep{Bishop:2006:PRM:1162264}. However, when there are many dimensions, most approaches from machine learning cannot be applied due to scalability issues. To circumvent this problem, we propose the use of the Conditional Variational Autoencoder (CVAE) \citep{KingmaCVAE, SohnCVAE}---a deep generative model which will be described in more detail below in Section~\ref{sec:cvae}. 

Another point worth noting is that the individuals $s_{0,i}$ can either come from the observed data or be sampled from another distribution $s_{0,i} \sim P(S|x_{t=0,i}, t=0)$. In the latter case, it is possible to generate larger and richer samples and this is a common approach when applied to population synthesis as it is a way to partly impute non-responses in the original data. However, modelling based on $P(S|x_{t=0}, t=0)$ will introduce additional noise and due to this we study travel preferences only for the observed individuals.

\begin{figure}[t!]
\centering
\includegraphics[height=7.0cm]{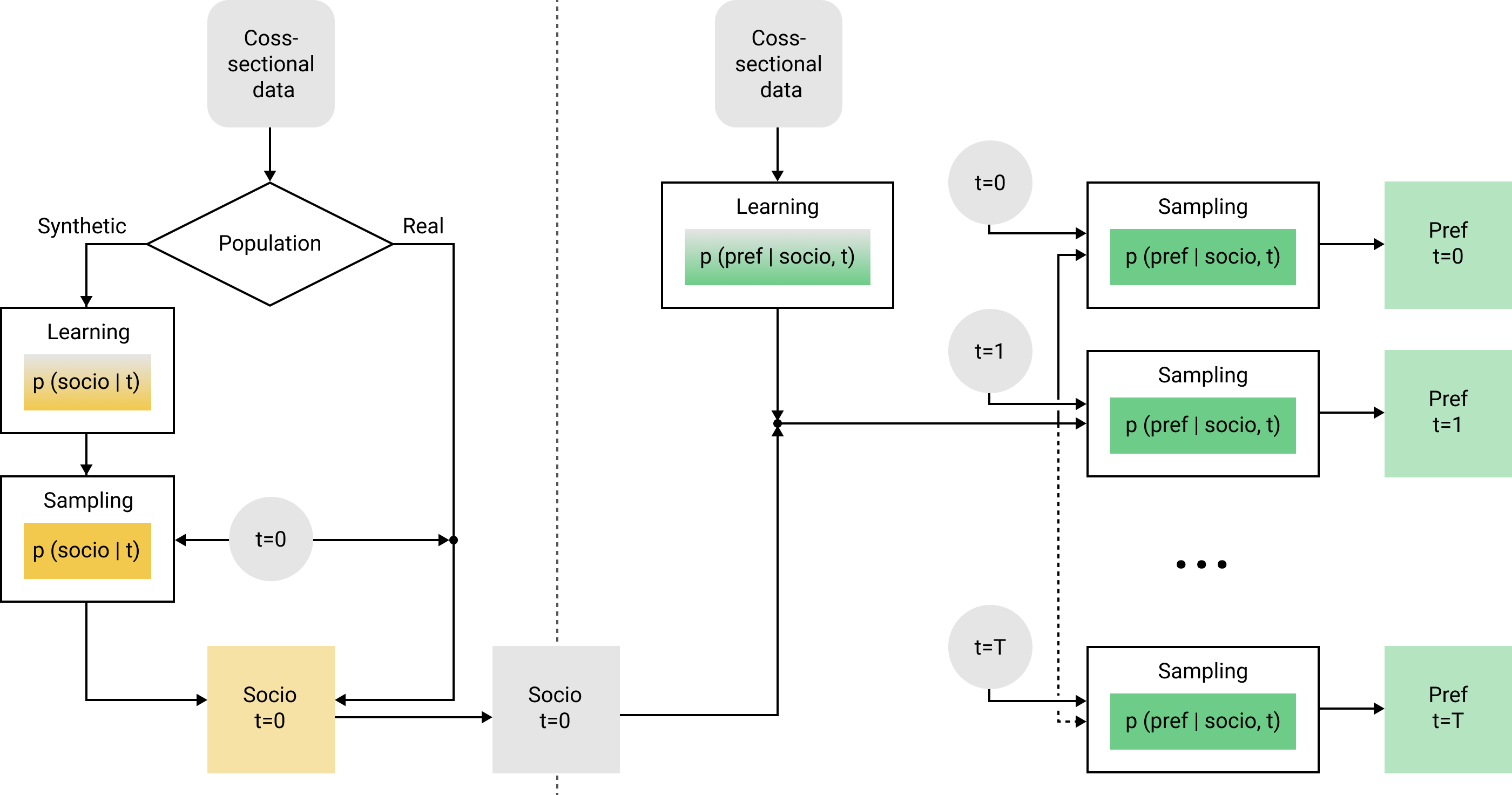}
\caption{Construction of a preference super pseudo-panel.}
\label{fig:pp}
\end{figure}

The framework for generating an SPP is depicted in Fig.~\ref{fig:pp} and can be summarized as follows. A CVAE model is used to 'learn' preference distributions from the travel diary data for all the years conditional on socio-economic and external attributes. Then, the pseudo panel can be created by sampling a number of the preference realizations for every year for a fixed pool of individuals from the reference year $t=0$, which in turn will form the corresponding preference distribution.

Insight into how the framework can be used from a model perspective may be better understood by considering the structure of a simple dynamic model. Although the data structure is not a panel and cannot be used to reveal causality and specific dynamic panel effects, it can 'fuel' a sophisticated pseudo-panel model for a given realization $r$ of a variable $Y$ of interest,
\begin{equation}
\label{eq:regr}
Y_{t, i, r} = \sum\limits_j \theta_j Z_{t, i, j} + \epsilon_{t, i} + \epsilon_{r|t, i}.
\end{equation}
In this model, $Z_{t, i, j}$ represents explanatory variables which may include the beforementioned attributes related to $V$, $S$ and $X$. $\epsilon_{t, i}$ represents unobserved noise related to the generation of $Y$ for individual $i$ at time $t$. In contrast to common econometric models, where $\epsilon_{t, i}$ represents unobserved elements not accounted for by the parametric model and noise accumulated in the data generating process, this term now, in addition, reflects prediction noise generated from the CVAE. However, as this noise is generally unsystematic and has the properties of 'white noise' which can be achieved from a proper specification and effective training,
it represents a modest sacrifice for being able to track individuals and not aggregated cohorts through time. The second error term $\epsilon_{r|t, i}$ is the variation across different realizations of preferences conditional on $t$ and individual $i$, representing different draws from the CVAE. The above model can be viewed as a random effect model in these domains and the suite of random effect estimators can be applied directly. In this paper, we will not estimate models on the basis of the synthetically generated data. However, this is indeed possible  although it will require the modeller to consider to what extent the CVAE model layer, between the original data and parametric model, will affect the precision of the model parameters. Insight into this can be supported by bootstrapping of the CVAE as considered in Section~\ref{sec:case:spp_construct}.      

%-------------------------------------------------------------------------

\section{Conditional Variational Autoencoder}
\label{sec:cvae}

As a scalable approach to model the high-dimensional joint distributions of preferences, we employ the Conditional Variational Autoencoder (CVAE) \citep{KingmaCVAE, SohnCVAE}. It represents a trivial extension of the Variational Autoencoder (VAE) \citep{kingma2013auto} which was previously used for the population synthesis problem in \cite{2018arXiv180806910B}. 

The main objective for the VAE, being a generative model, is the estimation of the joint distribution $P(V)$ of random variables $V$. In the CVAE, this objective is extended to include conditional variables $C$ (here we assume $C\equiv\{S, x_{t}, t\}$) and estimate $P(V \vert C)$ instead. However, as this extension is mathematically trivial and involves adding the conditioning on $C$, this paper only provides a description of the CVAE. Below, a high-level explanation of the main components of the CVAE model is provided---the autoencoding variational Bayes algorithm and the autoencoder \citep{Hinton504}. For more technical details, the reader can refer to the original papers by \cite{kingma2013auto, KingmaCVAE, SohnCVAE}.

The CVAE is a latent variable model. Introducing a latent variable $Z$ with a distribution $P(Z \vert C)$ leads to the factorization of the modelled distribution as
\begin{equation}
\label{eq:cvae:fact}
P(V \vert C) = \int P(V \vert C, Z)P(Z \vert C)dZ.
\end{equation}
To approximate the unknown distribution $P(V \vert C,Z)$, another distribution $Q_\theta(V \vert C,Z)$ with parameters $\theta$ is introduced. The objective is to minimize the difference between these distributions measured by the Kullback-Leibler (KL) divergence, $\mathcal{D}_\mathrm{KL}[Q_\theta(V \vert C,Z)\Vert P(V \vert C,Z)]$. After a few algebraic transformations, this divergence can be expressed as
\begin{equation}
\label{eq:cvae:obj}
\log P(V \vert C) - \mathcal{D}_\mathrm{KL}[Q_\theta(Z \vert V, C) \Vert P(Z \vert V, C)] = \mathbb{E}[\log P_\phi(V \vert Z, C)] - \mathcal{D}_\mathrm{KL}[Q_\theta(Z \vert V, C) \Vert P(Z \vert C)],
\end{equation}
where $P_\phi(V \vert Z, C)$ is the data distribution with parameters $\phi$ conditioned on the latent variables. The first term in the r.h.s. of Eq.~(\ref{eq:cvae:obj}), $\mathop{\mathbb{E}}[\log P_\phi(V \vert Z, C)]$, is a likelihood of the data projected from the latent space. The second term, $\mathcal{D}_\mathrm{KL}[Q_\theta(Z \vert V, C) \Vert P(Z \vert C)]$, is an error term corresponding to the projection of the data into the latent space. For modelling purposes, it is necessary to select a tractable form of $P(Z \vert C)$ so the second term can be efficiently calculated. A common choice is that of Gaussian distributions with zero mean and unit covariance matrix, i.e. $P(Z \vert C)=\mathcal{N}(\mathbf{0}, \mathbf{I}), \forall C$. Other choices, for example, more suitable for discrete distributions are also possible \citep{rolfe2016discrete,maddison2016concrete,jang2016categorical}, however, these are not considered in the current paper.

The autoencoding variational Bayes algorithm is a stochastic variational inference and learning algorithm based on maximizing the lower bound of $\log P(V \vert C)$ (the l.h.s. of Eq.~(\ref{eq:cvae:obj})), also known as evidence-lower bound (ELBO), by maximizing the r.h.s of Eq.~(\ref{eq:cvae:obj}) with respect to the observed data. This is based on maximum likelihood estimation of the parameters $\theta$ and $\phi$. $Q_\theta(V \vert C,Z)$ and $P_\phi(V \vert Z, C)$ represent functions that map the data to the latent space and back. In principle, these two functions can have an arbitrary form. However, the autoencoder model represents an effective and convenient choice.

The autoencoder is an artificial neural network that is specifically designed to find the latent space representation of the data. It has a characteristic symmetric shape consisting of two parts---the encoder and the decoder---mirrored with respect to each other. The encoder takes a data point and projects it to the latent space, while the decoder does the opposite. In the simplest case, both encoder and decoder can be fully connected feed-forward artificial neural networks (ANNs). For instance, assume that the encoder has four layers: an input layer of size $n$, two hidden layers with $A$ and $B$ neurons each, and an output layer that renders a latent representation vector of size $D_z$. In this case, the decoder also consists of four layers: an input layer of size $D_z$, two hidden layers with $B$ and $A$ neurons each, and an output layer of size $n$. 

In the following, we briefly describe how the a fully connected feed-forward ANN model processes the data. Each layer $l$ in the neural network takes the output from the previous layer $\mathbf{y}_{l-1}$ (where $\mathbf{y}_0$ is an input to the network), computes its weighted sum, $\mathbf{y}'_l=\mathbf{W}_l\mathbf{y}_{l-1}+\mathbf{b}_l$ (where $\mathbf{W}_l$ and $\mathbf{b}_l$ represent matrices of weights and biases for the layer) and applies a non-linear transformation $f$, $\mathbf{y}_l=f\left(\mathbf{y}'_l\right)$. A large number of choices for $f$ have been proposed. Commonly used functions include the hyperbolic tangent function, the logistic function and the rectified linear unit function, $\mathrm{ReLU}(x)=\max(0,x)$. The output neurons in the final layer that correspond to numerical variables usually have a linear form, while the soft-max function is usually used for categorical variables. During the training of the neural network, the difference between the output layer and the desired output, commonly referred to as the loss function $\mathcal{L}$, is calculated and minimized by adjusting the ANN parameters (weights and biases). The optimization is usually guided by the gradients of the loss function, which can be efficiently calculated using the chain differentiation rule commonly known as 'back-propagation'. 

\begin{figure}[t!]
\centering
\includegraphics[height=6.0cm]{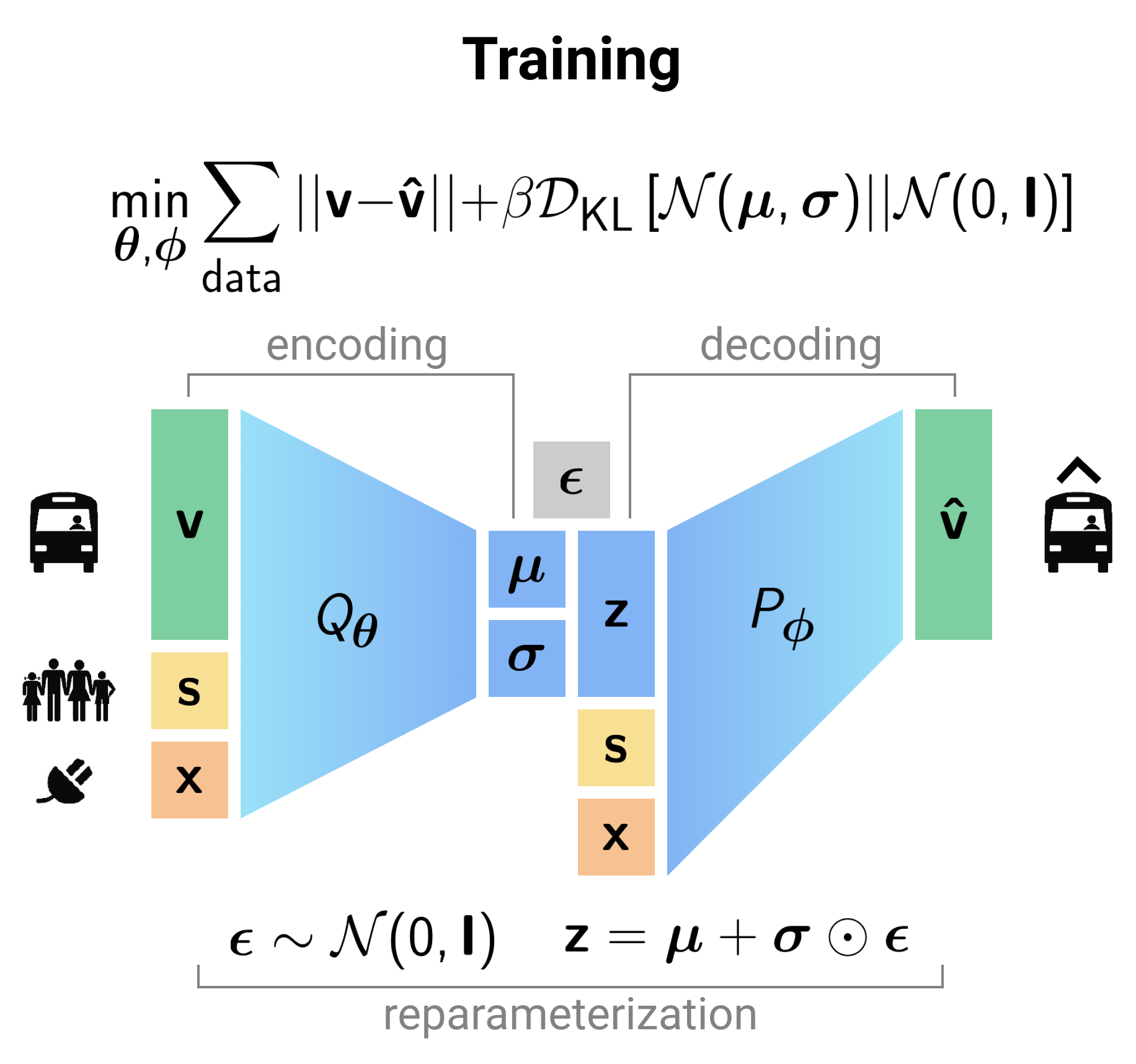}
\hspace{1.5cm}
\includegraphics[trim={2.0cm 0.0cm 2.0cm 0.0cm},clip,height=6.0cm]{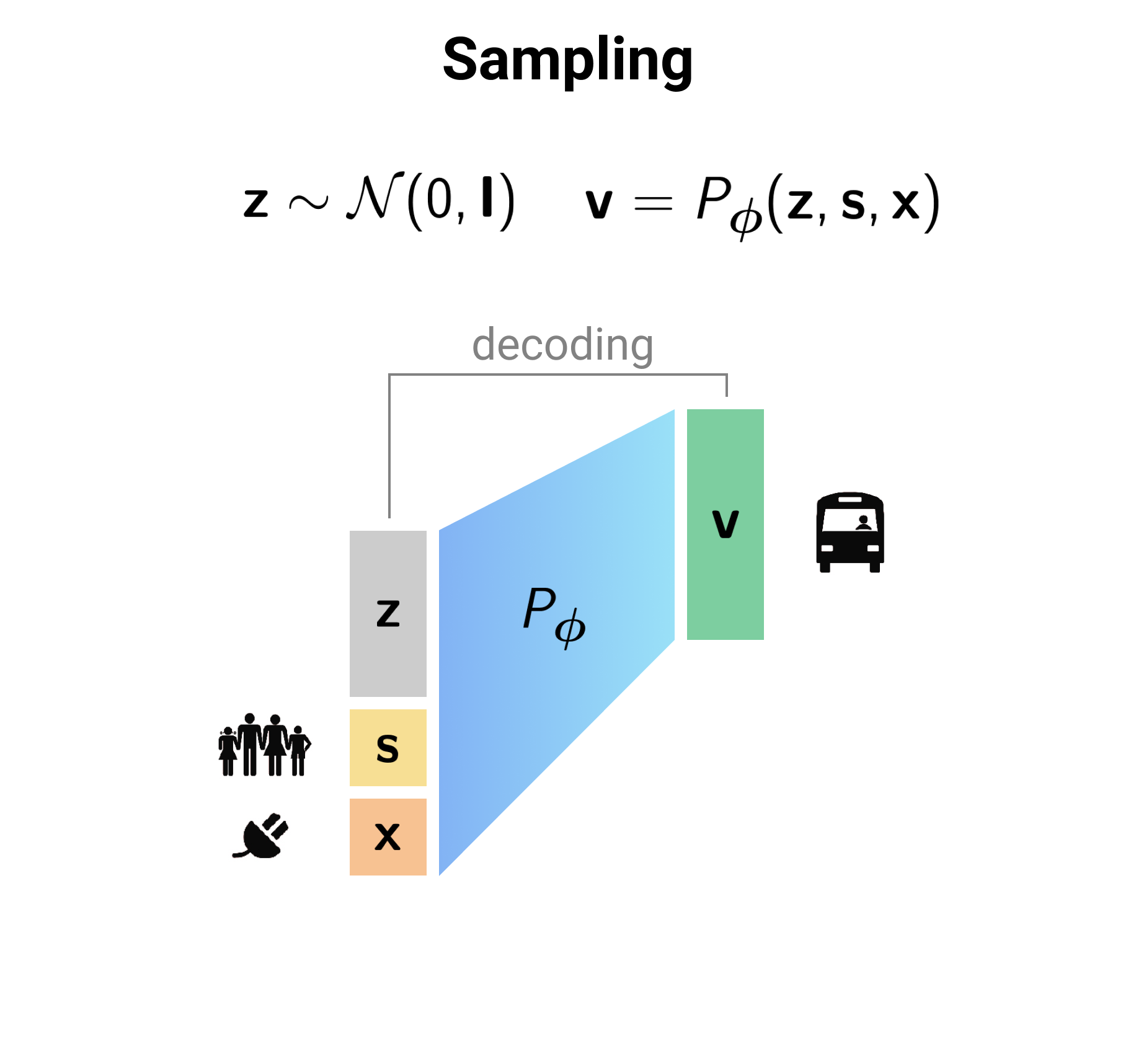}
\caption{Conditional Variational Autoencoder.}
\label{fig:cvae}
\end{figure}

As the 'correct' output from the autoencoder is its input itself, the training is performed in an unsupervised manner. In this case, the loss function is calculated as the difference between original data points and their reconstructed counterparts, i.e., it has the same form as the first term on the r.h.s. of Eq.~(\ref{eq:cvae:obj}). The loss is minimized by adjusting the encoder and decoder parameters, $\mathcal{W^\mathrm{e,d}}\equiv\left\{\mathbf{W}^\mathrm{e,d}_l, \mathbf{b}^\mathrm{e,d}_l\right\}_{l=1}^{N_l-1}$, where $N_l$ is the number of layers. Usually, the dimension of the latent space is chosen to be smaller than the original data ($D_z < n$) so the autoencoder has a butterfly shape. It leads to an information bottleneck in the latent space and forces the network to learn a compressed representation of the data as illustrated in Fig.~\ref{fig:cvae}. However, an alternative sparse version of the autoencoder with $D_z > n$ is also a possibility \citep{NIPS2006_3112}.

The architecture of the autoencoder is used as a 'backbone' of the CVAE model, where $Q_\theta(V \vert C,Z)$ and $P_\phi(V \vert Z, C)$ are modelled as the encoder and the decoder, respectively, with $\theta\equiv\mathcal{W^\mathrm{e}}$ and $\phi\equiv\mathcal{W^\mathrm{d}}$. The CVAE can also be interpreted as an  autoencoder with a probabilistic structure assigned to the latent space. Constraining the latent variables to follow some known distribution $P(Z \vert C)$ by the second term on the r.h.s. of Eq.~(\ref{eq:cvae:obj}) allows sampling from the model, where the decoder transforms known $P(Z \vert C)$ to $P_\phi(V \vert C)$ via a chain of nonlinear transformations represented by the neural network.

The way the CVAE processes the data is illustrated in Fig.~\ref{fig:cvae}. During the training (Fig.~\ref{fig:cvae}, left), the encoder $Q_\theta$ takes an input vector $\mathbf{v}_k$ concatenated with the conditional variables $\mathbf{s}_k$, $\mathbf{x}_k$ and $t_k$ ($k$ is the data sample index in the cross-sectional data used for the CVAE training) and outputs two vectors $\bm{\mu}_k$ and $\bm{\sigma}_k$ that correspond to the mean and the standard deviation of the Gaussian latent variable $Z$. Its realization $\mathbf{z}_k$ then has to be generated by sampling from $\mathcal{N}(\bm{\mu}_k,\bm{\sigma}_k)$. However, as the sampling operation is not differentiable, this makes it impossible to back-propagate the errors. To overcome this problem, $\mathbf{z}_k$ can be calculated indirectly as $\mathbf{z}_k=\bm{\mu}_k+\bm{\sigma}_k\odot\mathbf{\epsilon}_k$, where $\mathbf{\epsilon}_k\sim\mathcal{N}(\mathbf{0},\mathbf{I})$, and $\odot$ denotes element-wise multiplication. This approach is also known as the 'reparametrization trick' \citep{kingma2013auto}. Then, $\mathbf{z}_k$ is concatenated with the conditional variables $\mathbf{s}_k$, $\mathbf{x}_k$ and $t_k$ and passed through the decoder $P_\phi$ to generate a reconstructed version of the original data point, $\mathbf{\hat{v}}_k$. The output neurons in the decoder that correspond to numerical components of the data point $\mathbf{v}_k$, $v_{ik}$ ($i\in\{\mathrm{num}\}$) have a linear form, while the soft-max function, $\exp\left(v^{(m)}_{kj}\right)/\sum_{m=1}^{D_j}\exp\left(v^{(m)}_{kj}\right)$, is used for categorical components $v_{kj}$ ($j\in\{\mathrm{cat}\}$) taking one of the $D_j$ possible values within a ``one-hot'' representation. For example, the values in the $k^\mathrm{th}$ observation of a categorical variable $j$, which can take one of the $D_j=3$ possible values, can be encoded as (1, 0, 0), (0, 1, 0) and (0, 0, 1), i.e., $v^{(m)}_{ik}=1$ if the observation belongs to the category $m$ and $v^{(m)}_{ik}=0$ otherwise. Finally, the loss $\mathcal{L}(\theta,\phi)$, corresponding to the negative r.h.s. of Eq.~(\ref{eq:cvae:obj}), is calculated for all the data points in the training set and minimized with respect to the encoder and the decoder parameters,
\begin{equation}
\label{eq:loss}
\min\limits_{\theta,\phi}\mathcal{L(\theta,\phi)} = \sum\limits_k||\mathbf{v}_k-\mathbf{\hat{v}}_k||_\mathrm{num} + ||\mathbf{v}_k-\mathbf{\hat{v}}_k||_\mathrm{cat} + \beta \mathcal{D}_\mathrm{KL}\left[ \mathcal{N}(\boldsymbol{\mu}_k,\boldsymbol{\sigma}_k) || \mathcal{N}(0,\mathbf{I}) \right],
\end{equation}
where
\begin{equation}
\label{eq:loss_num}
%\begin{split}
||\mathbf{v}_k-\mathbf{\hat{v}}_k||_\mathrm{num} = \frac{1}{2}\sum\limits_{i\in \{\mathrm{num}\}} \left(v_{ki} - \hat{v}_{ki}\right)^2
\end{equation}
is the mean square loss associated with the reconstruction of numerical variables, whereas
\begin{equation}
\label{eq:loss_cat}
||\mathbf{v}_k-\mathbf{\hat{v}}_k||_\mathrm{cat} =  - \sum\limits_{j\in \{\mathrm{cat}\}}\sum\limits_{m=1}^{D_{j}} v^{(m)}_{kj}\log \hat{v}^{(m)}_{kj}
\end{equation}
is the cross-entropy loss associated with the reconstruction of categorical variables, taking one of the $D_j$ possible values within the one-hot representation, and
\begin{equation}
\label{eq:loss_KL}
\mathcal{D}_\mathrm{KL}\left[ \mathcal{N}(\boldsymbol{\mu}_k,\boldsymbol{\sigma}_k) || \mathcal{N}(\mathbf{0},\mathbf{I}) \right] = - \frac{1}{2}\sum\limits_{i=1}^{D_Z}\left(1 + \log\sigma_{ki} - \mu_{ki}^2 - \sigma_{ki}\right).
%\end{split}
\end{equation} 
is the KL divergence for two Gaussian distributions represented in a closed form. Here, we also use an additional weighting coefficient $\beta$ which controls the balance between the reconstruction error and influence of the prior. Varying this parameter can help the model to learn better latent representations of the data \citep{higgins2016beta}.

Having the CVAE trained, new data samples can be generated by sampling of the latent variable from the prior  $\mathcal{N}(\mathbf{0},\mathbf{I})$ and transforming it through the decoder (Fig.~\ref{fig:cvae}, right). It is worth noting that the exploration of the latent space will inevitably result in the generation of new samples that differ from those in the original data. For example, sampling $Z$ values on a line connecting two points in the latent space will result in the generation of samples characterized by a (non-linear) mixture of the attributes of the two samples generated from the line endpoints. However, these new samples are still valid in the sense that the correlations between their attributes are preserved.

%-------------------------------------------------------------------------

\section{Model validation}
\label{sec:model_eval}

In this section, we discuss evaluation methodologies for generative models used to approximate the conditional preference distribution $P(V \vert C)$. Following the standard procedure, the data is divided into training (held-in) and validation (held-out) sets. The training set is used to fit parameters of the model (weights and biases of the encoder and the decoder for the CVAE) whereas the model's performance is estimated on the validation set to demonstrate its ability to generalize. The generalization concept is closely related to the overfitting problem that occurs when the model fits the training data perfectly but performs poorly when applied to the validation set. The validation set is used to tune the hyper-parameters of the model, including the architecture of the encoder-decoder pair and the parameters of the optimization algorithm. The final model, which is found to perform the best on the validation set, is subsequently trained using the whole data set to maximize data utilization.

There are many ways to measure the model's performance on the validation set. The most straightforward approach is to directly use the ELBO value defined by Eq.~(\ref{eq:loss}). %This error consists of the data log-likelihood and the KL-divergence between the data posterior distribution in the latent space and the latent variable prior. 
Another approach is to compare statistical properties of the synthetic samples generated from the model to the statistical properties of the validation data using kernel density estimation. This can be represented by a multidimensional histogram in its simplest form. In this case, it requires calculation of the corresponding joint distributions $\hat{\pi}$ and $\pi$ for the two samples via simple cross-tabulation of counts for the attributes. Then, a distance measure between the two empirical distributions can be calculated. Here, we use the three common metrics while using the corresponding bin frequencies $\hat{\pi}_{i\dots j}$ and $\pi_{i\dots j}$: 
\begin{itemize}
\item Standardized root mean squared error (SRMSE),
\begin{equation}
\label{eq:srmse}
%\begin{split}
\mathrm{SRMSE} = \frac{\mathrm{RMSE}}{\overline{\pi}}
%= \frac{\sqrt{\overline{\left(\hat{p}-p\right)^2}}}{\overline{p}}
= \frac{\sqrt{\sum_{i}\dots\sum_{j}\left(\hat{\pi}_{i\dots j}-\pi_{i\dots j}\right)^2/N_\mathrm{b}}}{\sum_{i}\dots\sum_{j}\pi_{i\dots j}/N_\mathrm{b}},
%\end{split}
\end{equation}
where $N_\mathrm{b}$ is the the total number of bins compared;
\item Pearson correlation coefficient (Corr);
\item Coefficient of determination ($R^2$);
\end{itemize}
where the last two are defined in the standard way for the same bin frequencies. Other choices might involve, for instance, the KL-divergence or the Wasserstein metric, also known as the 'earth mover's' distance. Simplified approaches can also include comparison of different subsets of the distributions with respect to attributes and their marginals.

Finally, manual inspection of synthetic samples is also possible. This is a common approach in the case of human-readable formats such as images or text. For the tabular data studied in this paper, this approach has limitations. Although, in the simplest case, it might be based on different common sense rules, e.g. that a 10-year-old person cannot have a driving licence, the number of such rules grow exponentially with the number of attributes involved. However, manual inspection may still prove a valuable tool when analysing the performance of modelled  distributions of population cohorts. Moreover, as a means to inspecting the dynamic patterns of the panels and their fluctuations it is useful. As an example presented in the following subsection, Figure~\ref{fig:pref_cond_popreal_year_cond} illustrates quite clearly that dynamic effects based on the original data are affected by a substantial amount of sample noise. This becomes more pronounced when the number of dimensions increases. However, this effect is largely absent from the synthetic dynamic panel, as it rightly should be, which supports a 'cleaner' analysis of preferences. 

Other evaluation methods, which can be based on estimation of the regression performance of each variable using the rest of the variables (for instance, see \citep{Choi2017GeneratingMD}) or comparing clustering structures of the samples using, for example, principal component analysis, are not discussed in this paper.

Although the evaluation approaches described above work well for low-dimensional cases, when the number of modelled variables is small, they can often be misleading for large dimensions. They all suffer from the 'curse of dimensionality' problem, i.e., the number of samples required to cover the modelled space grows exponentially with the number of dimensions. Unfortunately, the most common approaches based on log-likelihood, kernel density estimates or visual fidelity of the samples fail to provide a 'silver bullet' solution and need to be application-specific \citep{theis2015note,Rainforth2018TighterVB}.

Our experiments confirm that tighter ELBO values do not correspond to better statistical properties of the synthetic samples and vice versa. Therefore, we use ELBO for tuning the optimization algorithm to stop training when the validation error approximately reaches its minimum, whereas the final model is selected based on the SRMSE error as it is more relevant for the application purposes. Moreover, the empirical distribution of the validation set becomes unreliable when it is estimated for all the variables due to the small number of samples available. To tackle this issue, we resort to the comparison of low-dimensional projections of the joint distributions on several subsets of the attributes as well as inspections of their marginals.

Using the statistical properties of the data as a means to select the best model can lead to an overfitting problem, which also manifests as low diversity of the generated samples. This problem especially becomes hard to detect when the statistical properties of the held-in data are very similar to the properties of the held-out data, and the model starts memorizing the held-in data to perfection. For instance, this is the case when a conventional Gibbs sampler is used to sample from the conditionals estimated as cross-tabulation of counts \citep{FAROOQ2013243,2018arXiv180806910B}. Although we use the early stopping and the bottleneck structure of the CVAE, where data are projected from the original high-dimensional space to the much more compact latent space, this itself cannot guarantee avoidance of the overfitting problem, and additional tests are thus required. The model should be able to generate new 'out-of-sample' data which are different from the original data but have similar statistical properties. To address this issue, we calculate overlaps between the generated samples and the training data by counting the percentage of samples which are the same in both data sets, where a 100\% overlap represents a case of pure overfitting.

Insight into the CVAE model performance can also be based on statistical bootstrapping. This technique is based on sampling with replacement, such that the size of the samples is equal to the size of the original data set. When applied to the training data, this enables the estimation of the variance of the statistics distribution for the modelled variables. For numerical variables, this can be based on the standard deviation of the mean, while either Shannon entropy or sum of squares can be used to estimate variance for categorical variables. This may also provide valuable insights with respect to the overall sample distribution and the model uncertainty in general and how this is dependent on different model specifications.

%-------------------------------------------------------------------------

\section{Case study: Transport preference dynamics in Denmark}
\label{sec:case}

In the remainder of the paper, the SPP methodology (Section~\ref{sec:frame}) and the CVAE model (Section~\ref{sec:cvae}) will be applied to study transport preference dynamics in Denmark, where these preferences are measured across a wide number of variables that describe transport activities in general. Using the SPP approach, it is possible to filter the preference changes that occur for these variables from one year to the next from the socio-economic variation resulting from either sampling noise or socio-economic trends. To the extent possible, the estimated preference changes are also filtered for changes to the infrastructure, including extensions to the road and public transport network. However, since accounting for these changes to the infrastructure at the local level over a period of 11 years is complicated, a simplified approach is used.

%-------------------------------------------------------------------------

\subsection{Data}
\label{sec:case:data}

The case study for generating a super pseudo panel of preferences is based on the Danish National Travel Survey (TU)\footnote{\url{http://www.modelcenter.transport.dtu.dk/english/tvu}} that represents a large continuous cross-sectional data set. It contains socio-economic characteristics of the participants, their geographic location as well as a detailed description of travel preferences throughout the day of the interview. 

Attribute variables used in the modelling are listed in Table~\ref{tab:data}. They can be divided into five groups:
\begin{enumerate}
    \item Year of the sample (attribute \#1), corresponding to the time variable $t$. We focus on the 11 subsequent years, from 2006 to 2016.
    \item Geographic location (attribute \#2), which is assumed to be an independent variable $g$. The modelled preference distribution is conditioned on $g$, i.e. $P(V \vert S, x_{t}, g, t)$.
    \item Changes in the infrastructure (attributes \#3--\#18), corresponding to the variables in $x_{t}$. To measure changes of infrastructure, level-of-service data from the Danish National Model \citep{richhansen2016} has been applied. These data exist in the form of level-of-service matrices for the respective modes and trip purposes. It is relatively complicated to maintain all network changes for all years. As a result, an interpolation approach \citep{richvandet2019} has been used where the level-of-service matrices from 2002, 2010 and 2015 are used to interpolate level-of-service matrices for all other years. Based on these matrices, we have calculated a number of level-of-service attributes at the level of the residential zone. These attributes are formulated as accessibility scores that define the number of jobs and people to be reached with car and public transport respectively within a set of predefined time intervals (10, 30, and 60 minutes). Clearly, this way of representing changes to the infrastructure does have limitations and in particular at the very local level. However, at the level of the population, this simplification is not likely to affect the results in a systematic way.
    \item Socio-economic characteristics (attributes \#19--\#32), such as a person's age, gender and income as well as a household's characteristics, corresponding to the variables in $S$.
    \item Travel preferences (attributes \#33--\#46), such as car ownership, distance travelled by different transport modes, number of trips reported, etc, corresponding to the variables in $V$. It should be noted that we do not attempt to model the complete trip diary. Firstly, because this is much more complicated as it involves numerous logical constraints with respect to the order and the timing of activities, and, secondly, because it is not relevant for the aim of the paper. As the paper is concerned with the dynamics of general transport preferences, it makes sense to look at a number of indicators as proposed.
\end{enumerate}

\begin{table}[t!]
\footnotesize
\centering
\begin{tabular}{lllll}
\hline
\# & Name & Number of values & Description \\
\hline
1 & DiaryYear & 11 & Year of the interview \\
2 & HomeAdrNTMzoneL2 & 868 & Home, zone in the Danish national transport model \\
\hline
3 & PersonScore10MinByCarL2 & 10 & Person accessibility score within 10 minutes by car \\
4 & PersonScore30MinByCarL2 & 10 & Person accessibility score within 30 minutes by car \\
5 & PersonScore60MinByCarL2 & 10 & Person accessibility score within 60 minutes by car \\
6 & PersonScoreAverageByCarL2 & 10 & Average person accessibility score by car \\
7 & PersonScore10MinByPubL2 & 10 & Person accessibility score within 10 minutes by public transport \\
8 & PersonScore30MinByPubL2 & 10 & Person accessibility score within 30 minutes by public transport \\
9 & PersonScore60MinByPubL2 & 10 & Person accessibility score within 60 minutes by public transport \\
10 & PersonScoreAverageByPubL2 & 10 & Average person accessibility score by public transport \\
11 & WorkScore10MinByCarL2 & 10 & Job accessibility score within 10 minutes by car \\
12 & WorkScore30MinByCarL2 & 10 & Job accessibility score within 30 minutes by car \\
13 & WorkScore60MinByCarL2 & 10 & Job accessibility score within 60 minutes by car \\
14 & WorkScoreAverageByCarL2 & 10 & Average job accessibility score by car \\
15 & WorkScore10MinByPubL2 & 10 & Job accessibility score within 10 minutes by public transport \\
16 & WorkScore30MinByPubL2 & 10 & Job accessibility score within 30 minutes by public transport \\
17 & WorkScore60MinByPubL2 & 10 & Job accessibility score within 60 minutes by public transport \\
18 & WorkScoreAverageByPubL2 & 10 & Average job accessibility score by public transport \\
\hline
19 & RespAgeCorrect & 8 & Age \\
20 & RespSex & 2 & Gender \\
21 & RespEdulevel & 11 & Educational attainment \\
22 & RespMainOccup & 12 & Principal occupation \\
23 & IncRespondent2000 & 10 & The respondent's gross income, price index 2000 \\
24 & HomeAdrCitySize & 8 & Home, town size \\
25 & HomeAdrDistNearestStation & 4 & Home, distance to nearest station \\
26 & HousehNumPers & 5 & Number of persons in the household \\
27 & HousehNumAdults & 5 & Number of adults (age $\geq$ 18) in the household \\
28 & IncHouseh2000 & 10 & The household's gross income, price index 2000 \\
29 & HousehAccomodation & 6 & Home, type \\
30 & HousehAccOwnorRent & 3 & Home, ownership \\
31 & NuclFamType & 4 & Nuclear family type \\
32 & PosInFamily & 4 & Position in the nuclear family \\
\hline
33 & RespHasBicycle & 2 & Bicycle ownership \\
34 & RespHasSeasonticket & 2 & Season ticket (public transport) \\
35 & ResphasDrivlic & 4 & Driver licence \\
36 & HousehNumcars & 5 & Number of cars in the household \\
37 & HousehNumDrivLic & 5 & Number of persons with driving licence in the household \\
38 & DayNumJourneys & 5 & Number of journeys during 24 hours \\
39 & NumTripsCorr & 4 & Number of trips \\
40 & PrimModeDay & 22 & Primary mode of transport for the entire day \\
41 & DayPrimTargetPurp & 27 & Primary stay of the day, purpose \\
42 & TotalLen & 5 & Total travel distance of trips \\
43 & TotalMotorLen & 5 & Total motorised travel distance \\
44 & TotalBicLen & 6 & Total bicycle travel distance \\
45 & TotalMin & 5 & Total duration of trips \\
46 & TotalMotorMin & 5 & Total motorised duration of trips \\
\hline
\end{tabular}
\caption{Variables from the cross-sectional data used in the paper: time (\#1), geographic location (\#2), changes in the infrastructure (\#3--\#18), socio-economic attributes (\#19--\#32), and travel preferences (\#33--\#46).}
\label{tab:data}
\end{table}

After removing the records with missing values, the data set contains 67419 records in total: 4345 (2006), 7010 (2007), 6606 (2008), 9885 (2009), 11966 (2010), 8354 (2011), 4783 (2012), 3600 (2013), 3541 (2014), 3172 (2015), and 4157 (2016). These data are used to train the CVAE model as discussed in the following subsection.

%-------------------------------------------------------------------------

\subsection{Model training and evaluation}
\label{sec:model_train}

The CVAE model is used to estimate the preference distribution $P(V \vert S, x_{t}, g, t)$. The data is divided into a training set (80\%), which is used to fit the CVAE parameters (weights and biases of the encoder and the decoder), and a validation set (20\%), which is used to compare performance of the CVAE models with different architectures and other hyper-parameters.

\begin{figure}[t!]
\centering
\includegraphics[height=5.0cm]{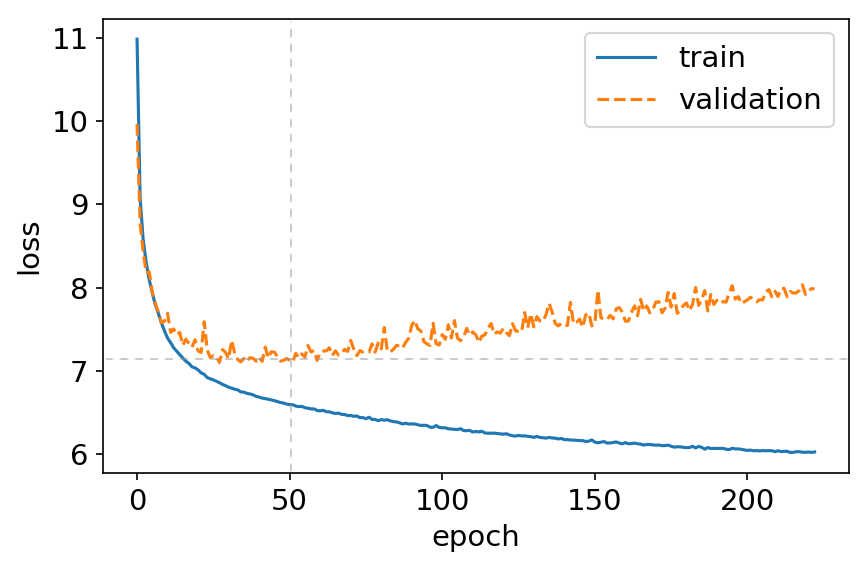}
\caption{Characteristic training and validation loss curves for the CVAE model. The training is stopped after 50 epochs, where the validation loss approximately has its minimum.}
\label{fig:cvae_loss_curves}
\end{figure}

The hyper-parameters are optimized using a standard grid search procedure. For the encoder, we explore a fully connected ANN architecture consisting of $N_l\in\{1, 2, 3\}$ hidden layers with $N_n / 2^l$ neurons each (integer division is assumed) with $tanh$ nonlinearity, where $l=0..N_l-1$ is the layer index and $N_n\in\{25, 50, 100, 200, 400\}$. The decoder has the same architecture of the encoder. The dimensionality of the latent space is $D_Z\in \{5, 10, 25\}$ and $\beta\in\{ 0.1, 0.5, 1.0, 10\}$. For the training, the RMSprop algorithm with a learning rate of 0.001 and $\rho=0.9$ is used. The size of a mini-batch is 64 and the number of epochs is 50. The latter is defined by the minimum of the loss (Eq.~(\ref{eq:loss})) calculated on the validation set (Fig.~\ref{fig:cvae_loss_curves}). All the numerical attributes are converted to categorical as this improved performance of the CVAE. For each numerical variable, the number of discrete bins used for the conversion is specified in the second column of Table~\ref{tab:data}. Thus, the cross-entropy loss (Eq.~(\ref{eq:loss_cat})) is applied for them instead of the mean square loss (Eq.~(\ref{eq:loss_num})).

The grid search found the best encoder architecture consisting of 3 hidden layers with 200, 100, and 50 neurons, respectively. The optimal hyper-parameters were found to be $\beta=0.5$, $D_z=5$. The best-performing model is subsequently trained using the whole data set to maximize data utilization.

\begin{figure}[t!]
\centering
\includegraphics[height=3.2cm]{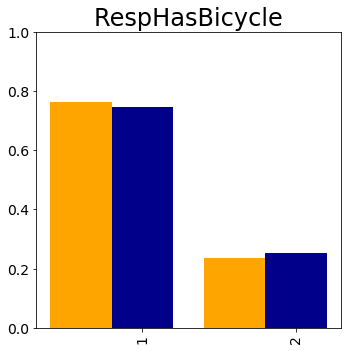}
\includegraphics[height=3.2cm]{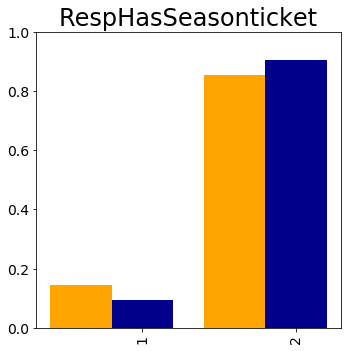}
\includegraphics[height=3.2cm]{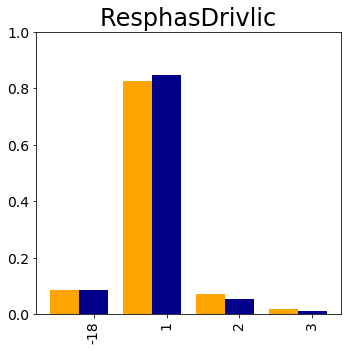}
\includegraphics[height=3.2cm]{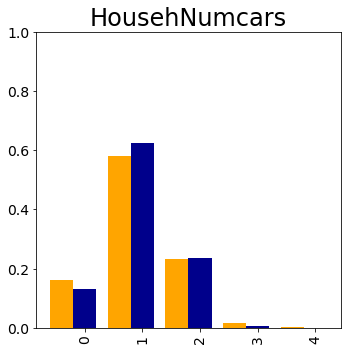}
\includegraphics[height=3.2cm]{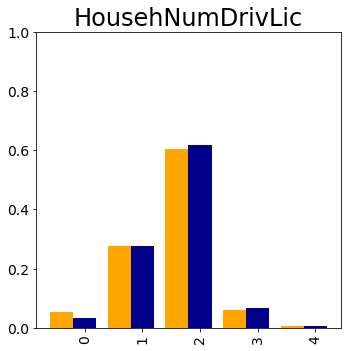}\\
\includegraphics[height=3.2cm]{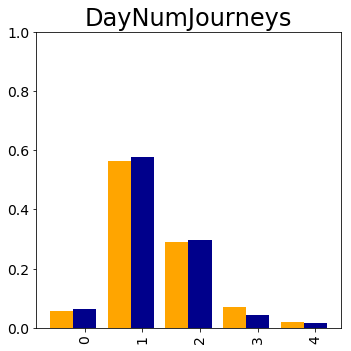}
\includegraphics[height=3.2cm]{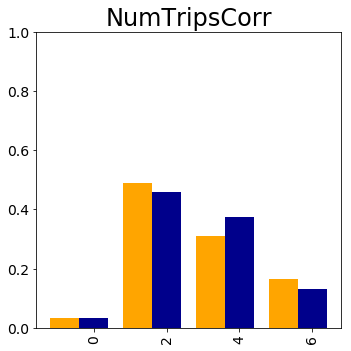}
\includegraphics[height=3.2cm]{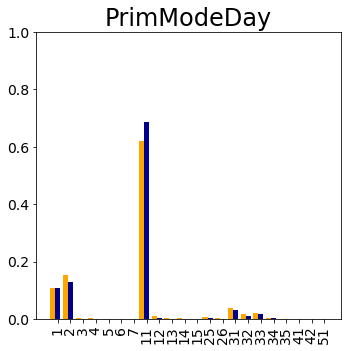}
\includegraphics[height=3.2cm]{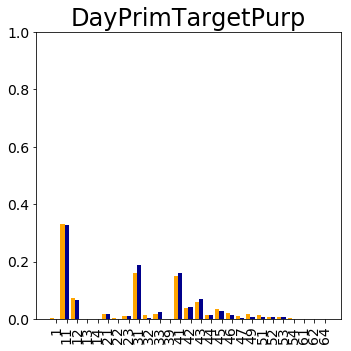}
\includegraphics[height=3.2cm]{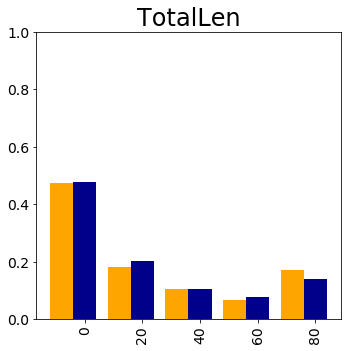}\\
\includegraphics[height=3.2cm]{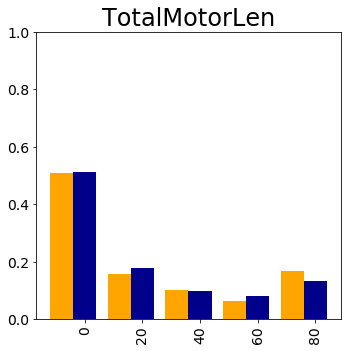}
\includegraphics[height=3.2cm]{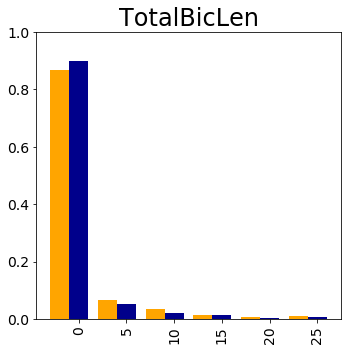}
\includegraphics[height=3.2cm]{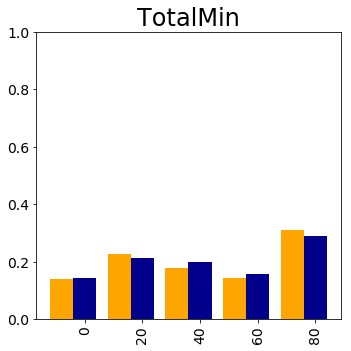}
\includegraphics[height=3.2cm]{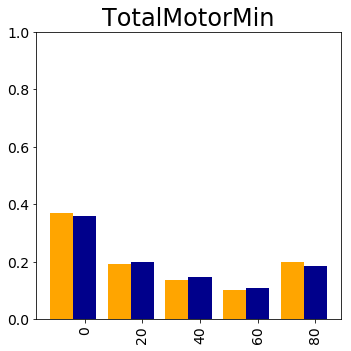}
\caption{Comparison of the marginal distributions of the preferences modelled by CVAE (dark blue) and the whole data set (orange). For the complete description of the categorical attributes, refer to the original TU documentation version 0616v1 at \protect\url{https://www.cta.man.dtu.dk/english/tvu/documentation}. Numerical attributes TotalLen, TotalMotorLen, and TotalBicLen are measured in kilometers; TotalMin and TotalMotorMin are measured in minutes.}
\label{fig:pref_marg}
\end{figure}

\begin{figure}[t!]
\centering
\hspace{0.5cm}(a)\hspace{3.8cm}(b)\hspace{3.8cm}(c)\hspace{3.8cm}(d)\\
\includegraphics[height=4.0cm]{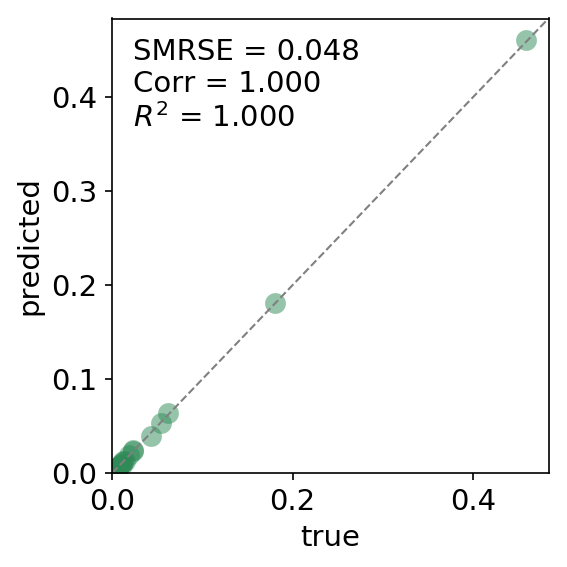}
\includegraphics[height=4.0cm]{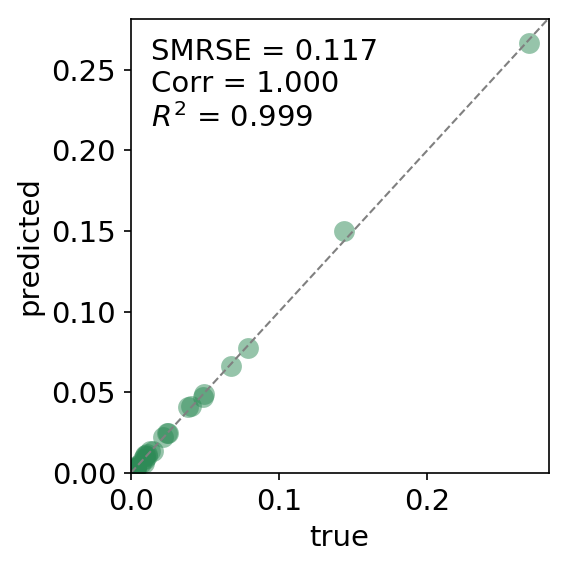}
\includegraphics[height=4.0cm]{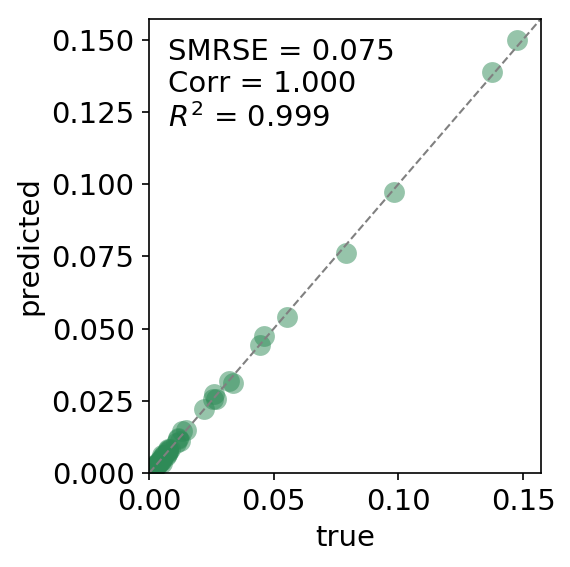}
\includegraphics[height=4.0cm]{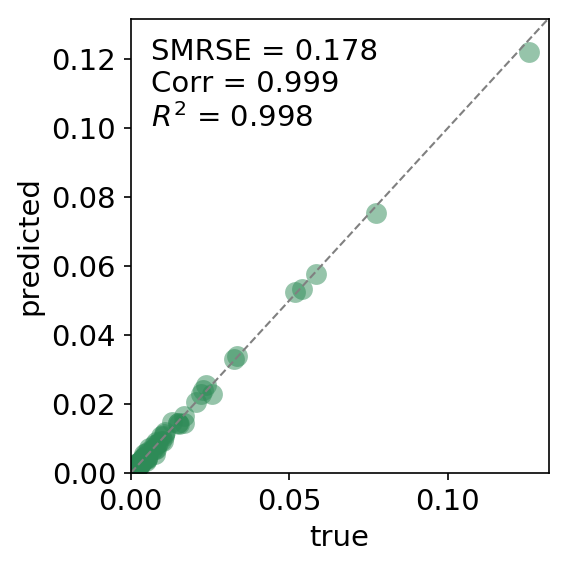}\\
\includegraphics[height=4.0cm]{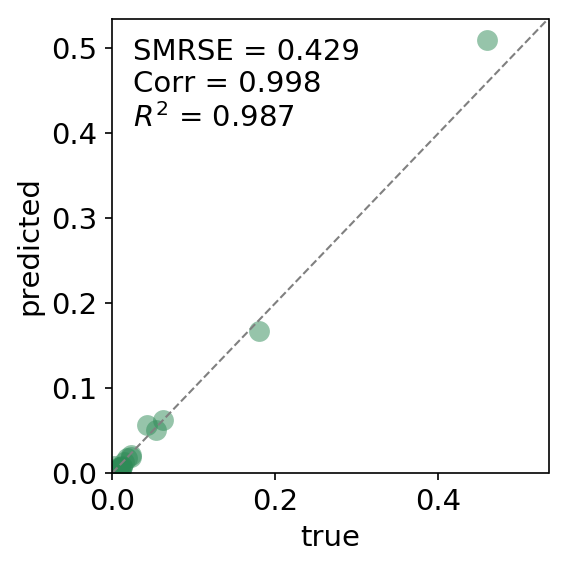}
\includegraphics[height=4.0cm]{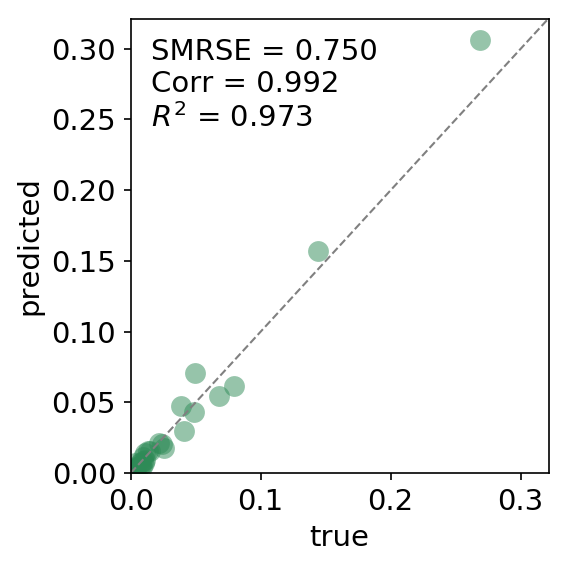}
\includegraphics[height=4.0cm]{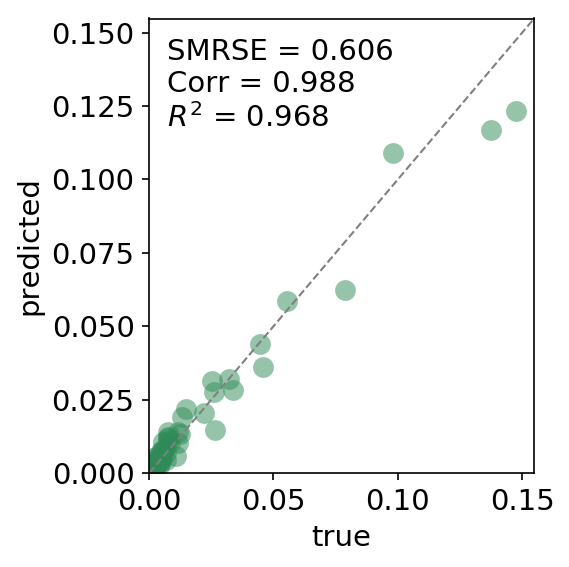}
\includegraphics[height=4.0cm]{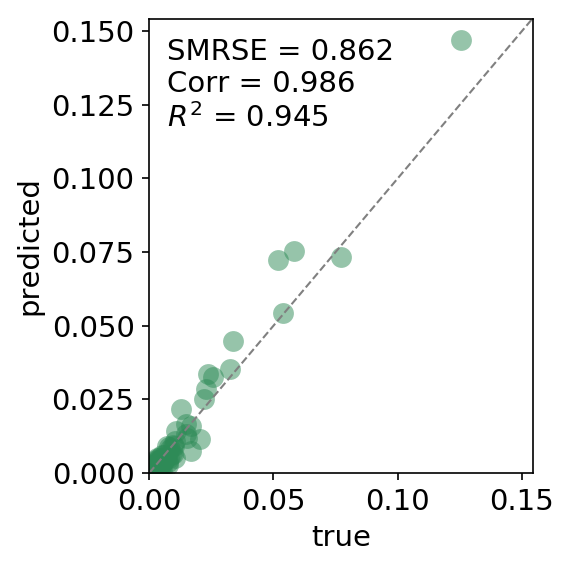}\\
\includegraphics[height=4.0cm]{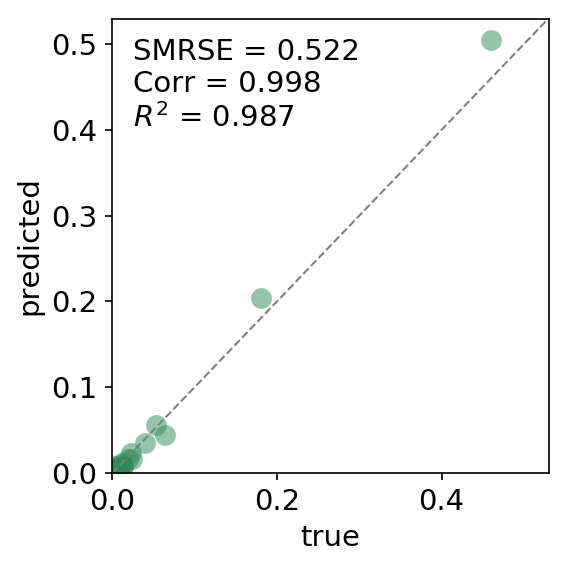}
\includegraphics[height=4.0cm]{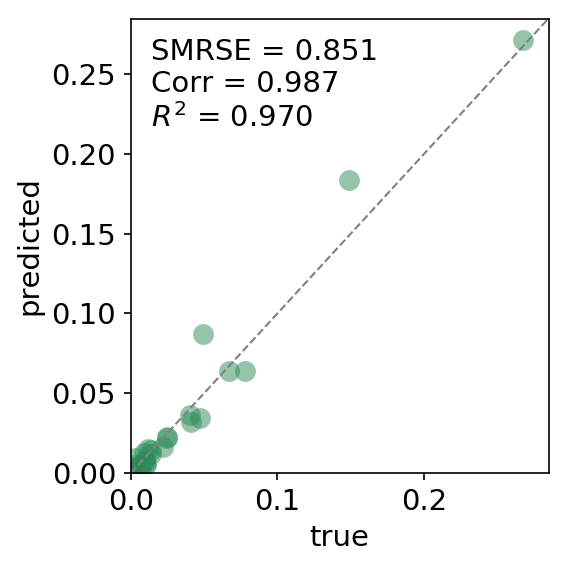}
\includegraphics[height=4.0cm]{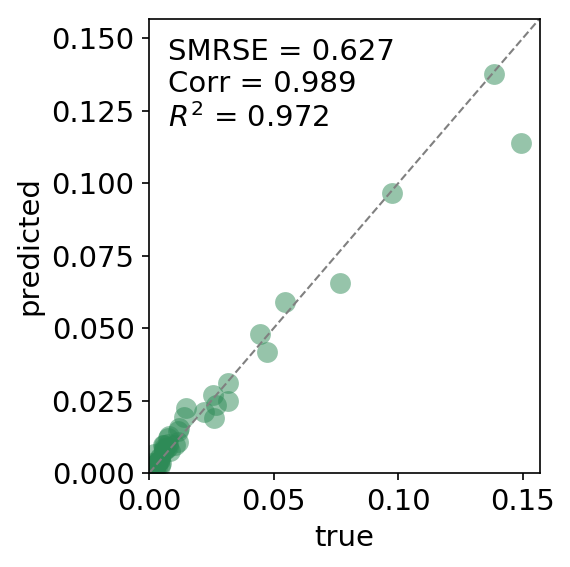}
\includegraphics[height=4.0cm]{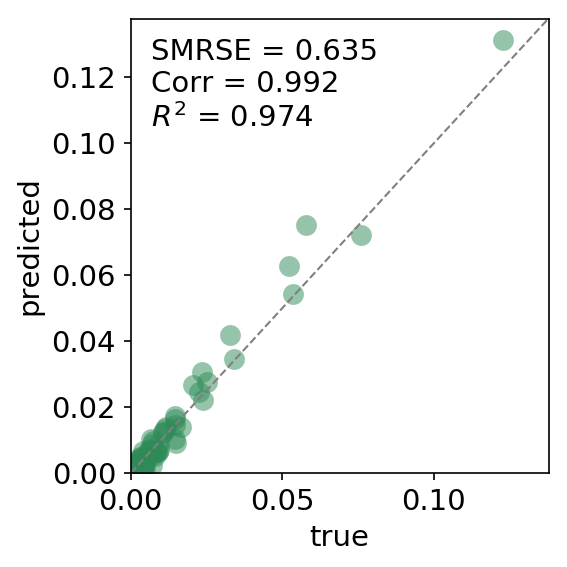}
\caption{Comparison of the preferences' joint distributions: Training set versus validation set (first row), samples from CVAE versus validation set (second row), and samples from the final CVAE and the whole data set (third row) projected onto four different subsets of attributes: RespHasBicycle $\times$ RespHasSeasonticket $\times$ ResphasDrivlic $\times$ TotalBicLen (first column), PrimModeDay $\times$ NumTripsCorr $\times$ DayNumJourneys (second column), TotalLen $\times$ TotalMotorLen $\times$ TotalMin $\times$ TotalMotorMin (third column), and DayPrimTargetPurp $\times$ HousehNumcars $\times$ HousehNumDrivLic (fourth column). See Table~\ref{tab:data_comparison} for more details about the subsets of attributes.}
\label{fig:pref_scatter}
\end{figure}

To evaluate the quality of the modelled preference distribution, we generate 100,000 synthetic samples using the CVAE and compare their statistical properties to the properties of the observed data. As discussed in Section~\ref{sec:model_eval}, we resort to the comparison of the low-dimensional projections of the joint distributions on several subsets of attributes as defined in Table~\ref{tab:data_comparison} as well as visual inspections of the marginals. Figure~\ref{fig:pref_scatter} shows the bin frequencies of the two empirical joint distributions $\hat{\pi}$ ('predicted') and $\pi$ ('true') plotted against each other. In the first row of the figure, $\hat{\pi}$ is estimated directly from the training data and $\pi$ is estimated from the validation data. It suggests that the statistical properties of the training and validation data are highly similar (largest SRMSE=0.178, smallest Corr=0.999 and $R^2$=0.998). The plots in the second row verify that the best-performing CVAE model shows a good fit to the validation data (largest SRMSE=0.862, smallest Corr=0.986 and $R^2$=0.945). Finally, the third row shows the performance of the best-performing CVAE model when it is trained on the whole data set and compared with the whole data set itself (largest SRMSE=0.851, smallest Corr=0.987 and $R^2$=0.970). Marginal distributions for the final CVAE model are depicted in Fig.~\ref{fig:pref_marg} and also suggest that the model fits the data well.

\begin{table}[t!]
\footnotesize
\centering
\begin{tabular}{llr}
\hline
Subset & Variables & Number of combinations \\
\hline
1.1 & RespSex $\times$ RespMainOccup $\times$ RespAgeCorrect & $2 \times 12 \times 8 = 192$ \\
1.2 & RespEdulevel $\times$ IncRespondent2000 & $11 \times 10 = 110$ \\
1.3 & HomeAdrCitySize $\times$ HomeAdrDistNearestStation & $8 \times 4 = 32$ \\
1.4 & NuclFamType $\times$ PosInFamily $\times$ HousehAccomodation $\times$ HousehAccOwnorRent & $4 \times 4 \times 6 \times 3 = 288$ \\
1.5 & IncHouseh2000 $\times$ HousehNumPers $\times$ HousehNumAdults & $10 \times 5 \times 5 = 250$ \\
\hline
2.1 & RespHasBicycle $\times$ RespHasSeasonticket $\times$ ResphasDrivlic $\times$ TotalBicLen & $2 \times 2 \times 4 \times 6 = 96$ \\
2.2 & PrimModeDay $\times$ NumTripsCorr $\times$ DayNumJourneys & $22 \times 4 \times 5 = 440$ \\
2.3 & TotalLen $\times$ TotalMotorLen $\times$ TotalMin $\times$ TotalMotorMin & $5 \times 5 \times 5 \times 5 = 625$ \\
2.4 & DayPrimTargetPurp $\times$ HousehNumcars $\times$ HousehNumDrivLic & $27 \times 5 \times 5 = 675$ \\
\hline
\end{tabular}
\caption{Subsets of attributes used for comparison of the joint distributions of the real samples and synthetic samples generated by CVAE. Subsets 1.1--1.5 and 2.1--2.4 are used for the population synthesis and preference synthesis problems, respectively.}
\label{tab:data_comparison}
\end{table}
 
To estimate diversity of the generated samples, we calculate overlaps between the generated samples and the training data by counting percentage of the samples that are the same in both data sets, which is 100\% in the case of pure over-fitting. The baseline overlap between the training and validation data sets is around 50\%, meaning that half of the samples in both data sets are different. The overlap of the best-performing CVAE and the training data set is around 62\% whereas the overlap of the best-performing CVAE and the validation data set is around 45\%. The overlap of the samples from the final CVAE trained on the whole data set and the whole data set itself is around 63\%, meaning that the remaining 37\% of the generated travel preference combinations are completely new, i.e., not present in the recorded data.

%-------------------------------------------------------------------------

\subsection{Analysis of travel preferences using SPP}
\label{sec:case:spp_construct}

Having verified that the CVAE has properly learned $P(V \vert S, x_{t}, g, t)$, we move to the construction of the SPP. Since the 4345 observations for 2006 are only around half of the observations in 2007 (7010), we use the 7010 individuals from 2007 as a base population $s_0$. This population, for which the socio-economic profiles and geographic locations $g_i$ are known, is now moved forward to the years 2008--2016 and backward to 2006. For each individual $i$, we sample travel preferences 1000 times for each year $t$ to numerically estimate the joint distribution $P_{t,i}(V)$. The sampled distributions also take into account changes in the infrastructure through the conditioning on the respective accessibility scores $x_{t,i}$ for each year.

\begin{figure}[t!]
\centering
(a)\hspace{3.6cm}(b)\hspace{3.6cm}(c)\hspace{3.6cm}(d)\\
\includegraphics[height=4.0cm]{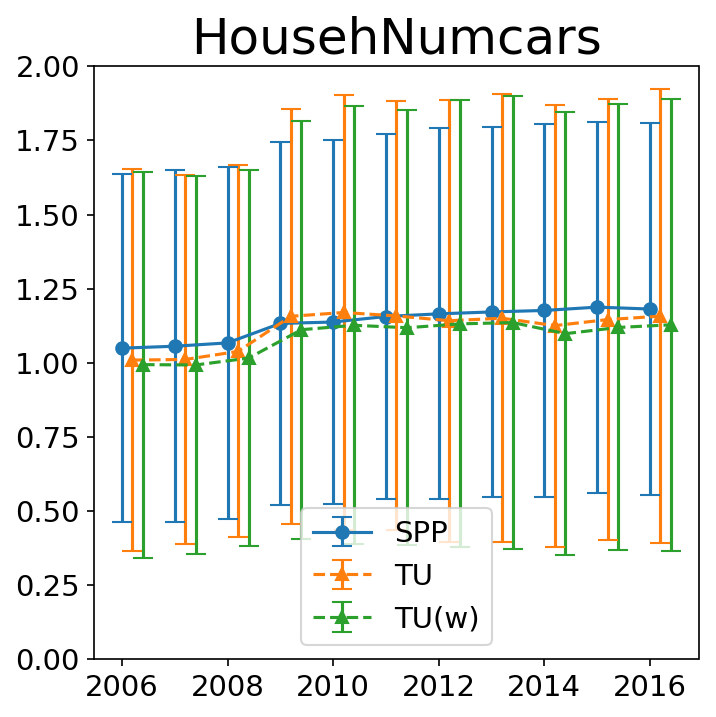}
\includegraphics[height=4.0cm]{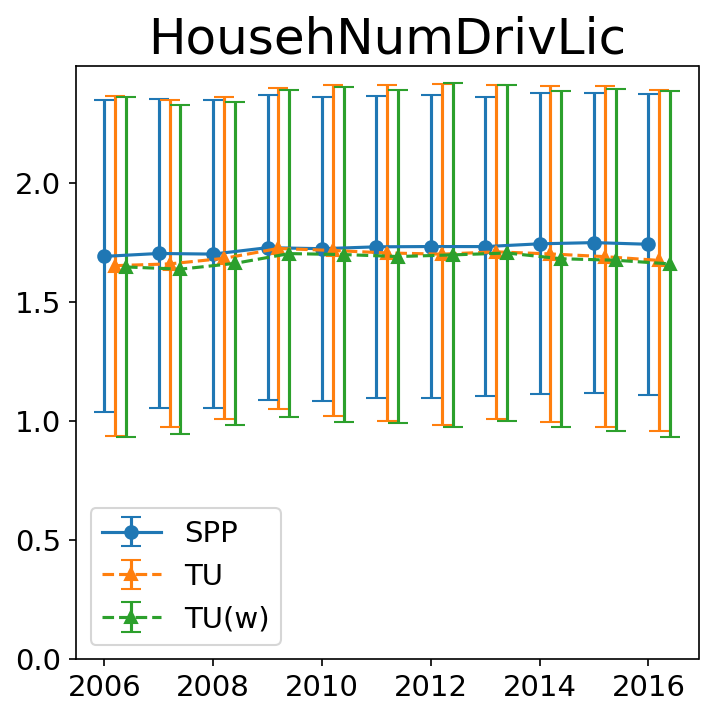}
\includegraphics[height=4.0cm]{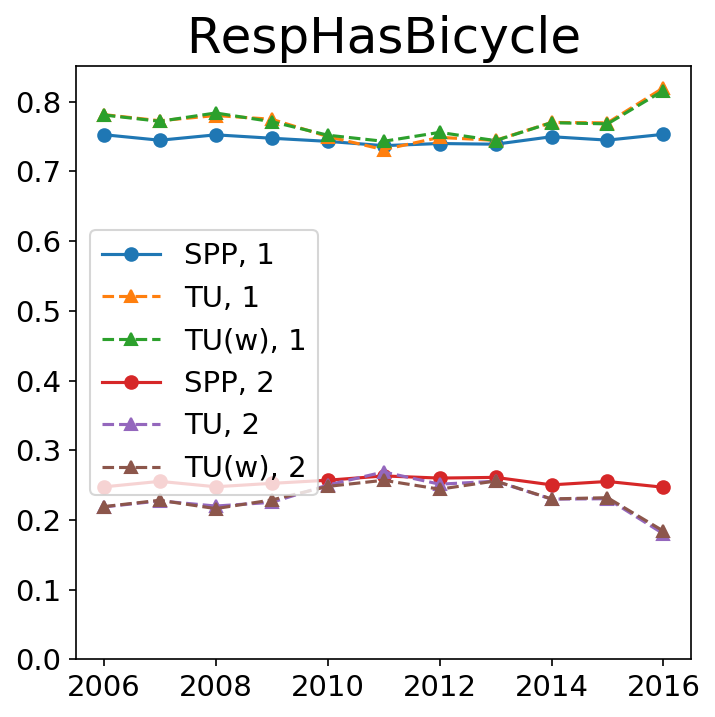}
\includegraphics[height=4.0cm]{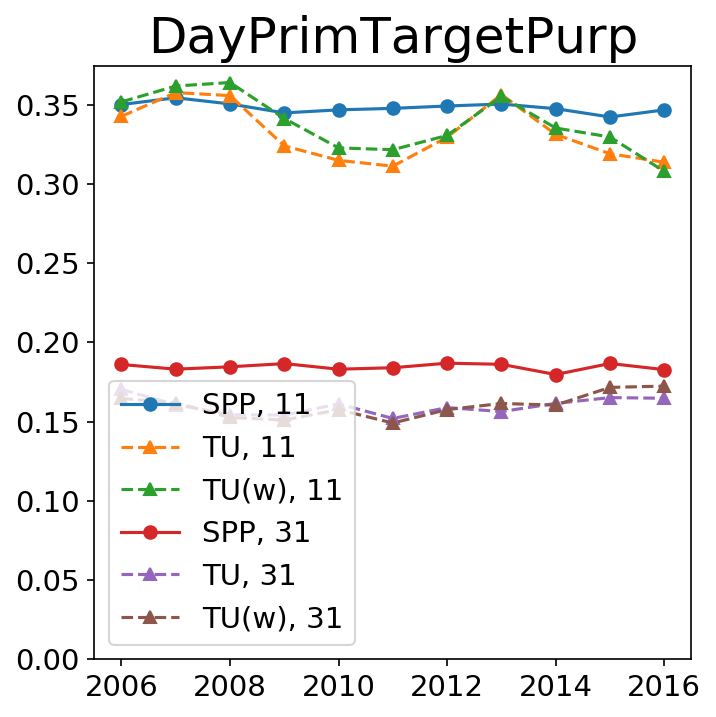}
\caption{Annual dynamics of two numerical (a, b) and two categorical (c, d) attributes in the SPP generated by CVAE (solid lines) for the fixed population of 2007, TU and weighted TU (dashed lines) with different population for each year. For the numerical attributes, their mean and standard deviation are plotted. For the categorical attributes, the probabilities of the most frequent categories are shown. For the description of the categorical variables, see Table~\ref{tab:pref_cat_desc}.}
\label{fig:pref_cond_popreal_year}
\end{figure}

Examples of preference dynamics for the whole population are shown in Fig.~\ref{fig:pref_cond_popreal_year}. For comparison purposes, we also plot a weighted version of the TU data, where sample weights are calculated to match aggregated national population statistics for a few selected socio-demographic attributes. In general, the modelled attributes indicate a close relationship with the cross-sectional data but less volatile due to the elimination of sample noise. For example, the growing trend in car ownership (Fig.~\ref{fig:pref_cond_popreal_year}(a)) is clearly captured as well as its increase in 2009, which is due to a technical change in the survey. The trends for the number of persons holding a driving licence in the household (Fig.~\ref{fig:pref_cond_popreal_year}(b)) and bicycle ownership (Fig.~\ref{fig:pref_cond_popreal_year}(c)) also show an acceptable model fit although in recent years divergence is observed. In some cases, the trend of the SPP is slightly different from that of the raw cross-sectional data. For example, the trends concerning the main trip purpose are quite different (Fig.~\ref{fig:pref_cond_popreal_year}(d)) in that distinct 'dips' for the observed commuter trips between 2008 and 2013 and later between 2013 and 2016 are noticed. It is generally acknowledged that travel preferences are relative stable over time and it is therefore quite unlikely that the volatile behaviour of the observed time series represents a general trend. It rather represents sample noise for these particular years, which may as well have been caused by the financial crisis. It should be noted that the impact of the financial crisis, to the extent it affects the employment status of individuals, is also filtered for in the SPP.

\begin{table}[t!]
\footnotesize
\centering
\begin{tabular}{lllll}
\hline
Attribute & Values & Description \\
\hline
RespHasBicycle & 1 & Yes \\
 & 2 & No \\
\hline
DayPrimTargetPurp & 11 & Workplace (normal workplace/address of employer) \\
 & 12 & School, educational institution \\
 & 31 & Shopping \\
 & 41 & Visit family/friends \\
 & 44 & Summer cottage, allotment \\
 & 45 & Walk, run, bicycle trip, drive (the trip was a purpose in itself) \\
\hline
\end{tabular}
\caption{Description of two categorical attributes. For the detailed description of the rest of the attributes, refer to the original TU documentation version 0616v1 at \protect\url{https://www.cta.man.dtu.dk/english/tvu/documentation}.}
\label{tab:pref_cat_desc}
\end{table}

\begin{figure}[t!]
\centering
(a1)\hspace{3.6cm}(a2)\hspace{3.6cm}(b1)\hspace{3.6cm}(b2)\\
\includegraphics[height=4.0cm]{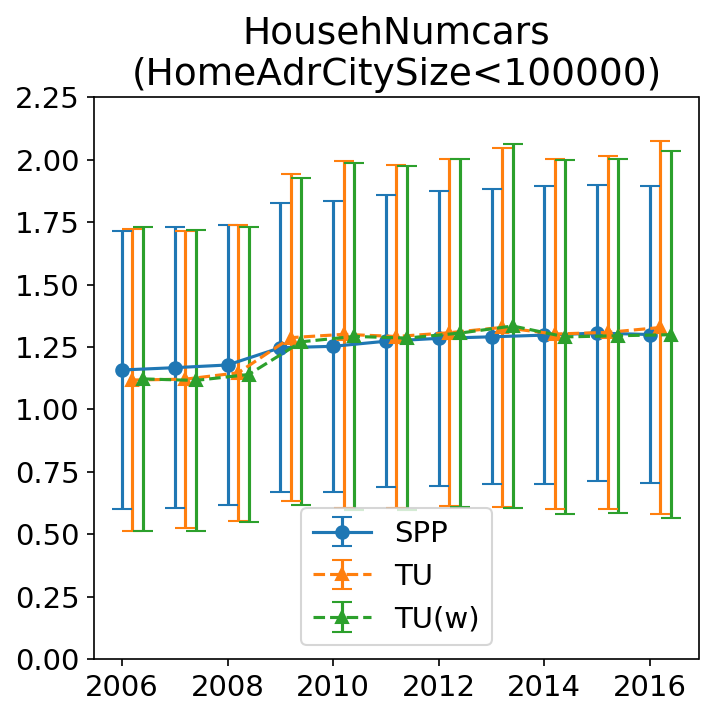}
\includegraphics[height=4.0cm]{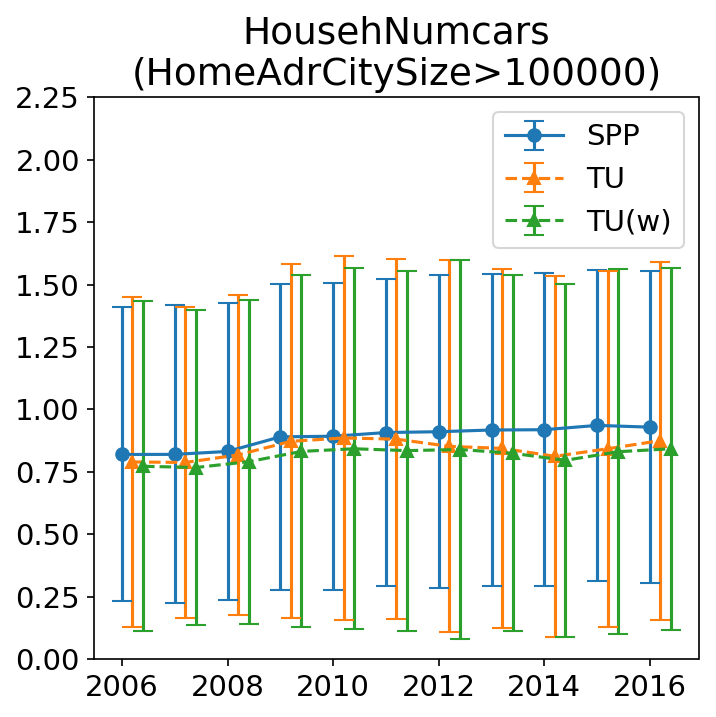}
\includegraphics[height=4.0cm]{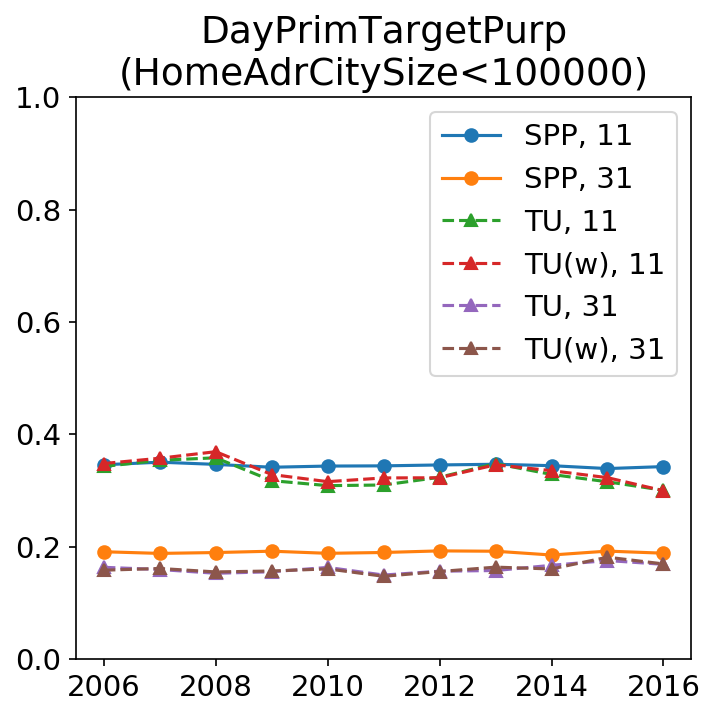}
\includegraphics[height=4.0cm]{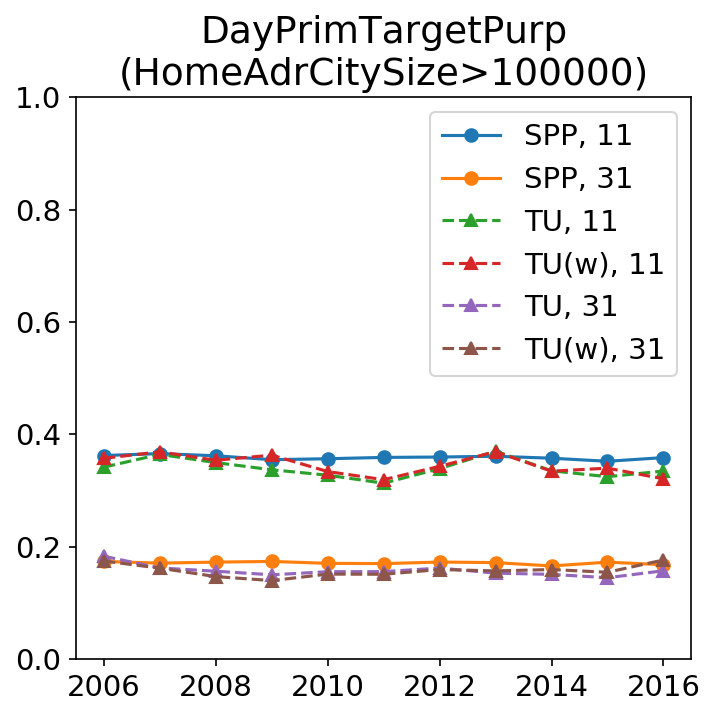}\\
(c1)\hspace{3.6cm}(c2)\hspace{3.6cm}(d1)\hspace{3.6cm}(d2)\\
\includegraphics[height=4.0cm]{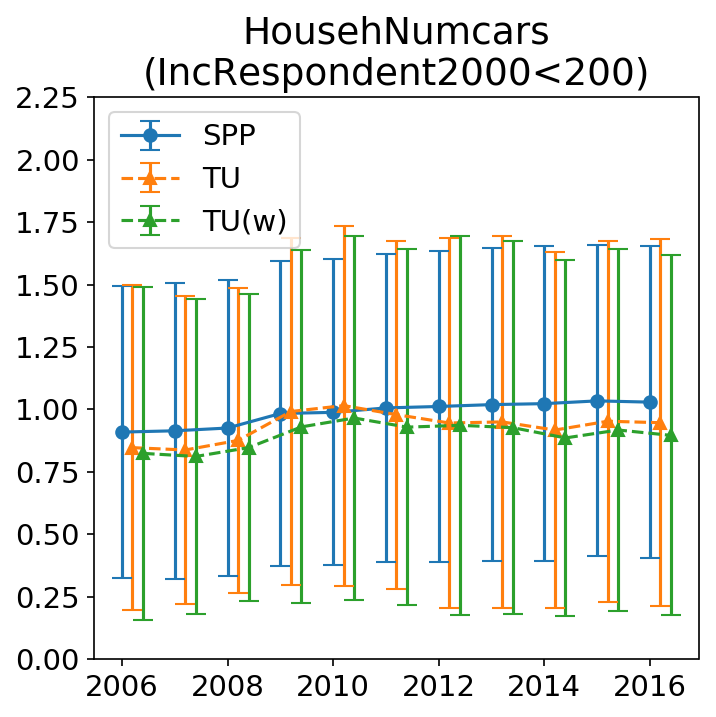}
\includegraphics[height=4.0cm]{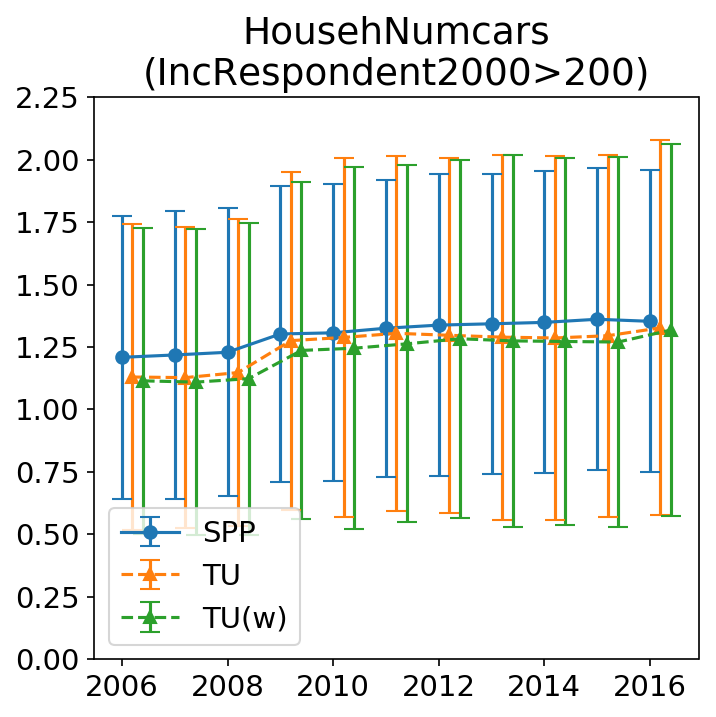}
\includegraphics[height=4.0cm]{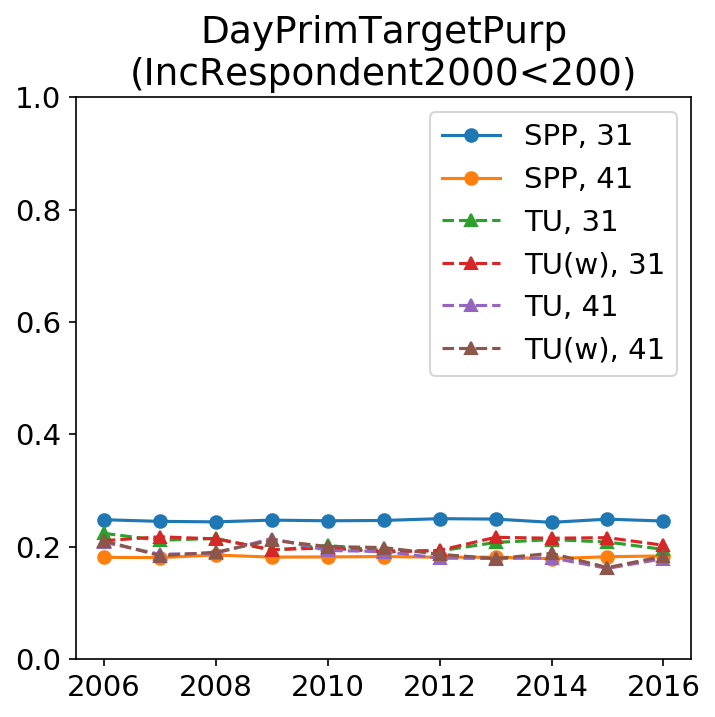}
\includegraphics[height=4.0cm]{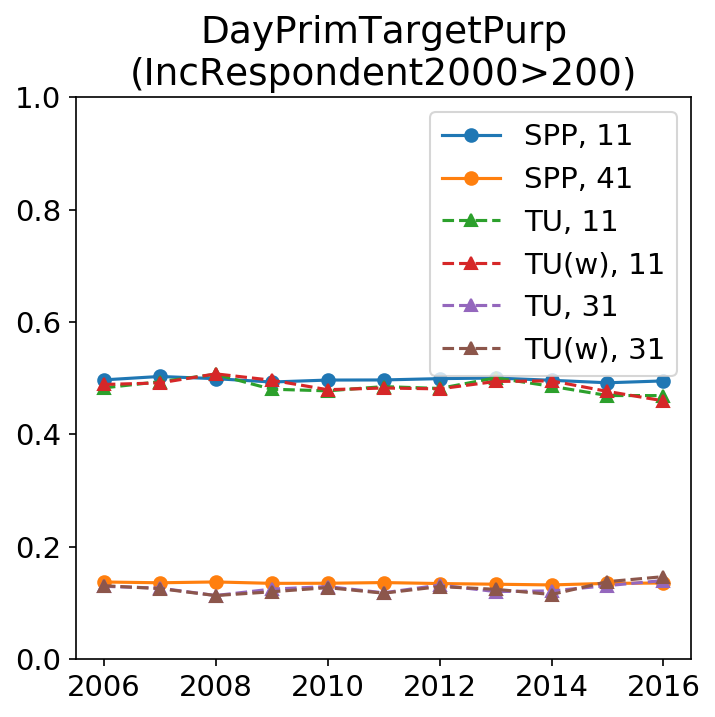}\\
(e1)\hspace{3.6cm}(e2)\hspace{3.6cm}(f1)\hspace{3.6cm}(f2)\\
\includegraphics[height=4.0cm]{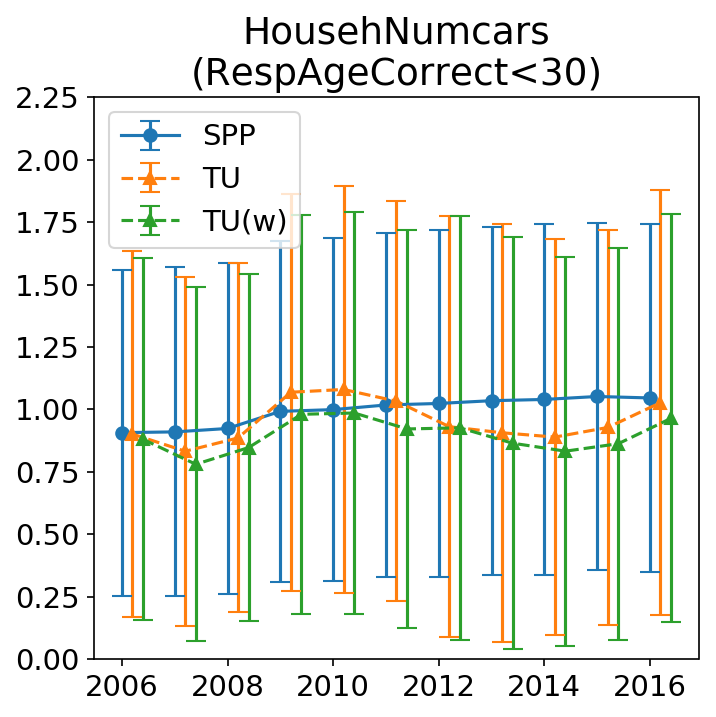}
\includegraphics[height=4.0cm]{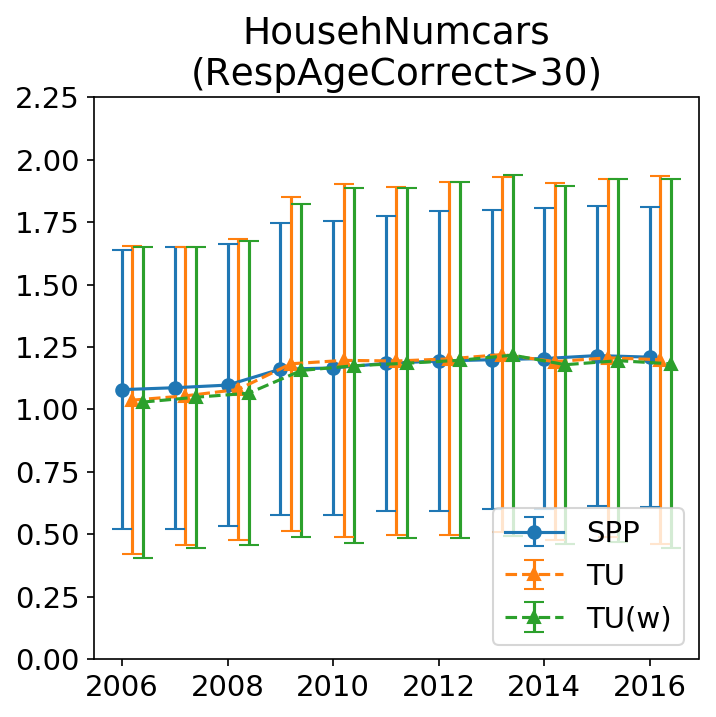}
\includegraphics[height=4.0cm]{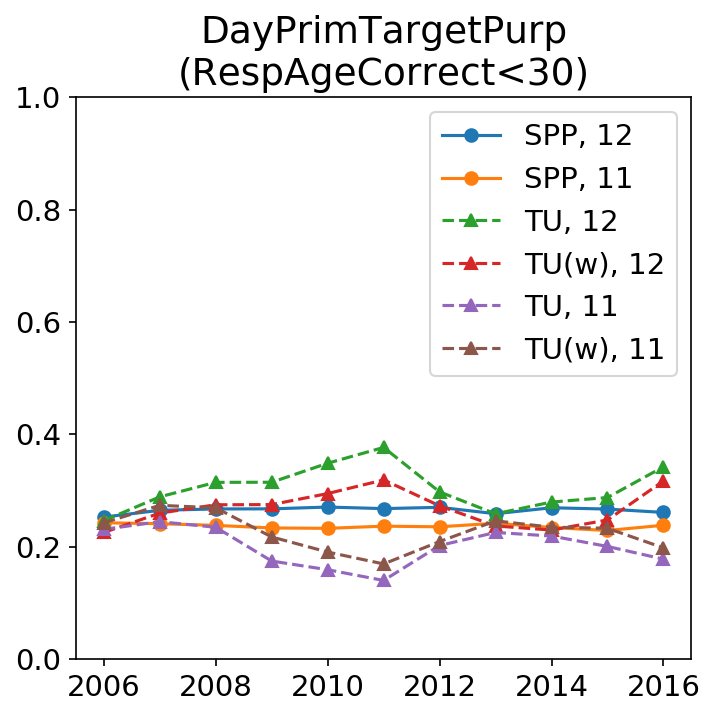}
\includegraphics[height=4.0cm]{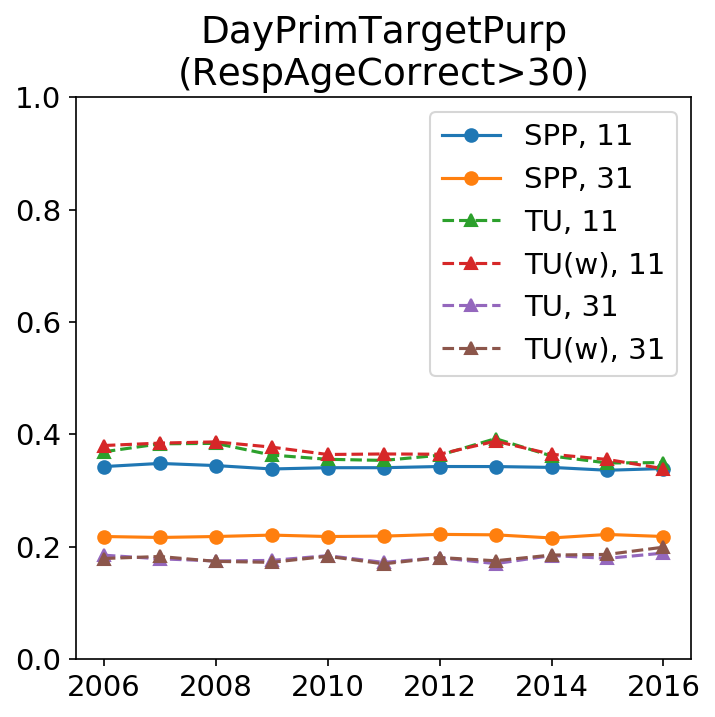}\\
(g1)\hspace{3.6cm}(g2)\hspace{3.6cm}(h1)\hspace{3.6cm}(h2)\\
\includegraphics[height=4.0cm]{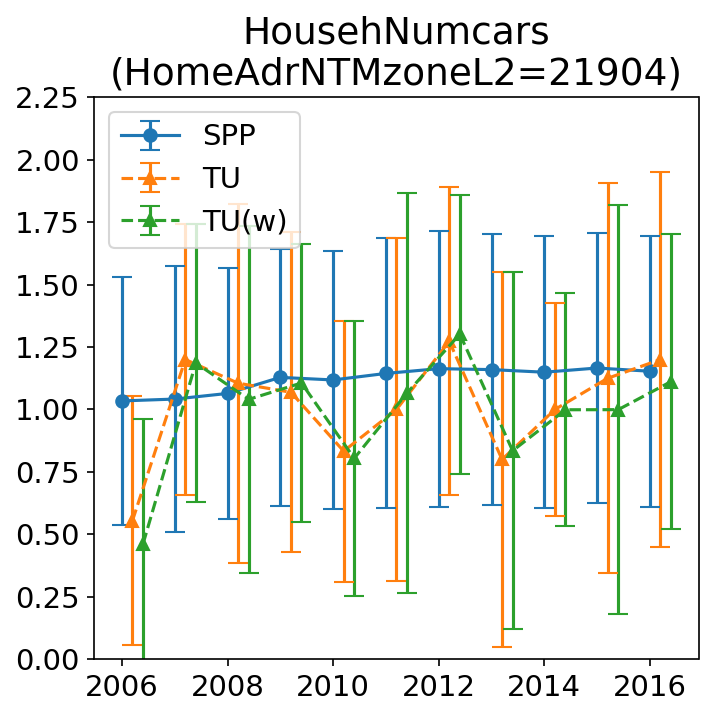}
\includegraphics[height=4.0cm]{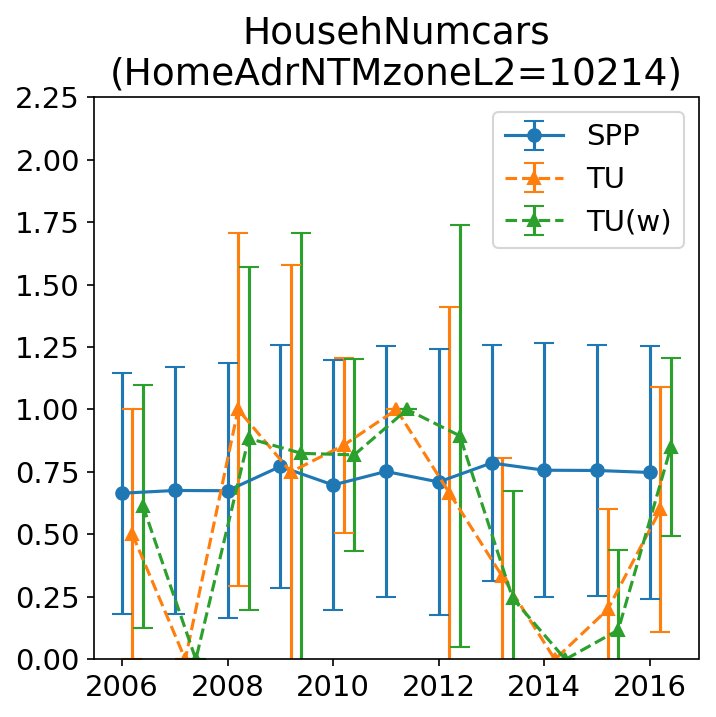}
\includegraphics[height=4.0cm]{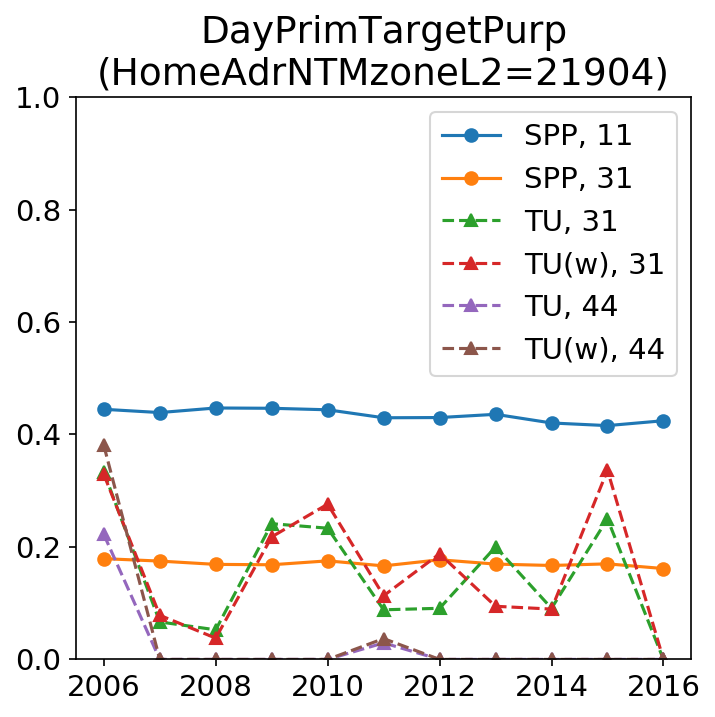}
\includegraphics[height=4.0cm]{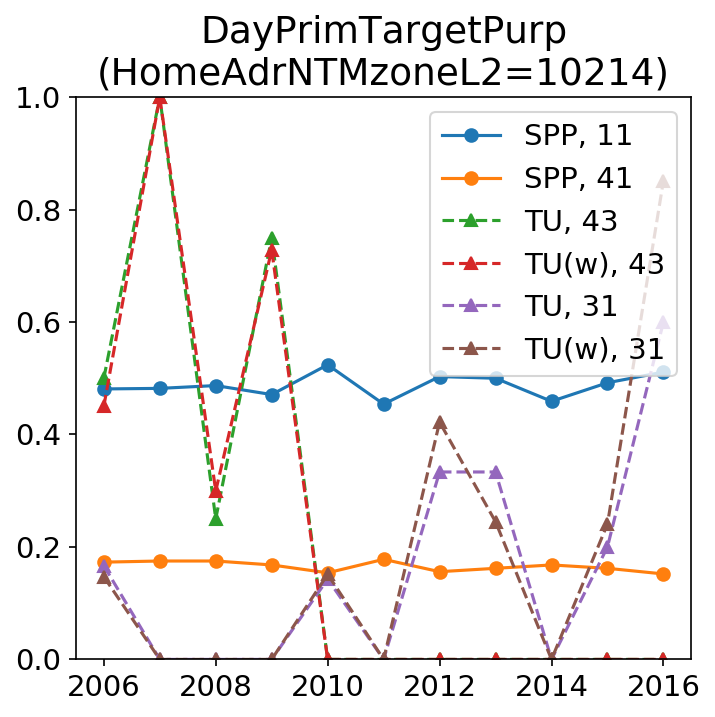}\\
\caption{Annual dynamics of one numerical (first column) and one categorical (second column) attributes in the SPP generated by CVAE (solid lines) for the fixed population of 2007, TU and weighted TU (dashed lines) with different population for each year conditional on HomeAdrCitySize (first row), IncRespondent2000 (second row), RespAgeCorrect (third row), and HomeAdrNTMzoneL2 (last row). For the numerical attribute, its mean and standard deviation are plotted. For the categorical attribute, the probabilities of the most frequent categories are shown. Sparsity of the TU data leads to significant fluctuations over the years while statistical properties of the SPP data remain more stable. For the description of the categorical variables, see Table~\ref{tab:pref_cat_desc}.}
\label{fig:pref_cond_popreal_year_cond}
\end{figure}

The constructed SPP allows for the comparison of travel preferences for specific groups of people. As an example, Figs.~\ref{fig:pref_cond_popreal_year_cond}(a,b) compare the car ownership and trip purpose dynamics for people living in rural and urban areas. Fig.~\ref{fig:pref_cond_popreal_year_cond}(c,d) and Fig.~\ref{fig:pref_cond_popreal_year_cond}(e,f) show dynamics of the same attributes for low versus middle/high income respondents and respondents below/above 30 years old, respectively. In all these three cases, the distinction between these groups of people is clearly captured. It is interesting to note that the car ownership dynamics depicted in Fig.~\ref{fig:pref_cond_popreal_year_cond}(e) is in almost perfect agreement for the population older that 30 years, while the oscillating behaviour for the young population in the TU data differs from the steadily growing trend revealed by the SPP. The benefits of jointly estimating the preference distribution for all years and across all combinations of socio-economic variables (including the geographical locations of the households) makes it possible to investigate preferences for single individuals and very detailed cohorts without introducing aggregation bias. For example, it is possible to investigate the relationship between education level and income or between age and trip purpose. Because of this fact, the SPP analysis can provide new insights with respect to how preferences change over time and at a very detailed level, and even if real data are locally scarce. For example, it is possible to compare populations from different geographical zones (Fig.~\ref{fig:pref_cond_popreal_year_cond}(g,h)), where the differences between a rural area near Hillerod (zone 21904) and an urban area in central Copenhagen (zone 10214) are clearly captured, whereas the TU data clearly suffers from severe sampling noise.

\begin{figure}[t!]
\centering
(a)\hspace{3.6cm}(b)\hspace{3.6cm}(b1)\hspace{3.6cm}(b2)\\
\includegraphics[height=4.0cm]{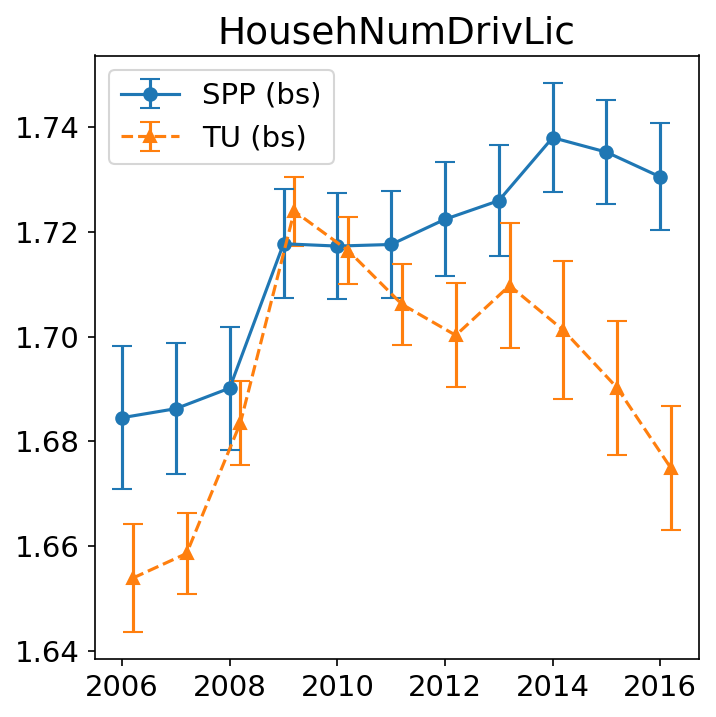}
\includegraphics[height=4.0cm]{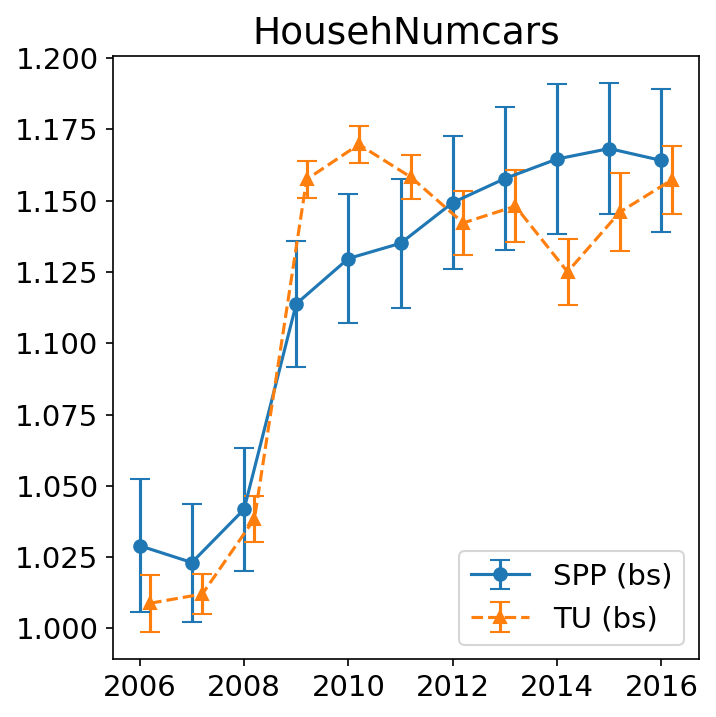}
\includegraphics[height=4.0cm]{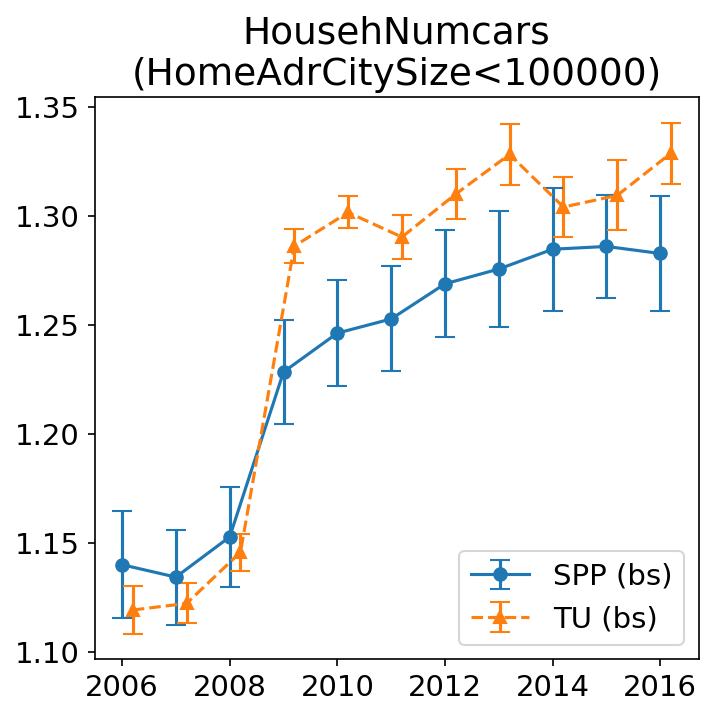}
\includegraphics[height=4.0cm]{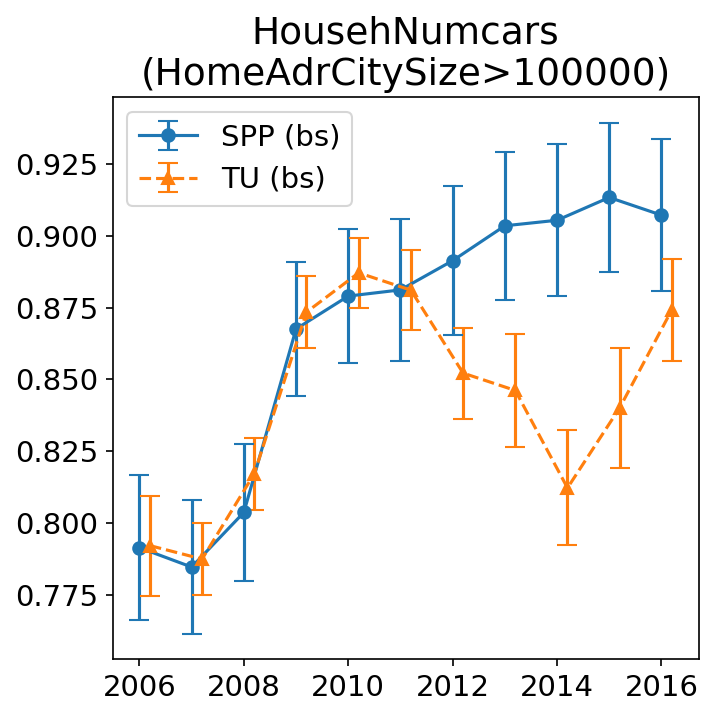}\\
(c1)\hspace{3.6cm}(c2)\hspace{3.6cm}(d1)\hspace{3.6cm}(d2)\\
\includegraphics[height=4.0cm]{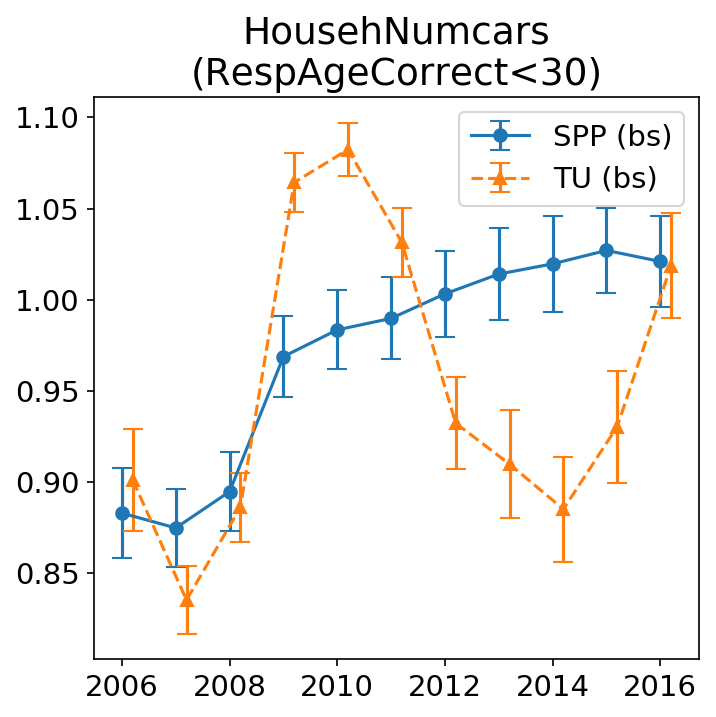}
\includegraphics[height=4.0cm]{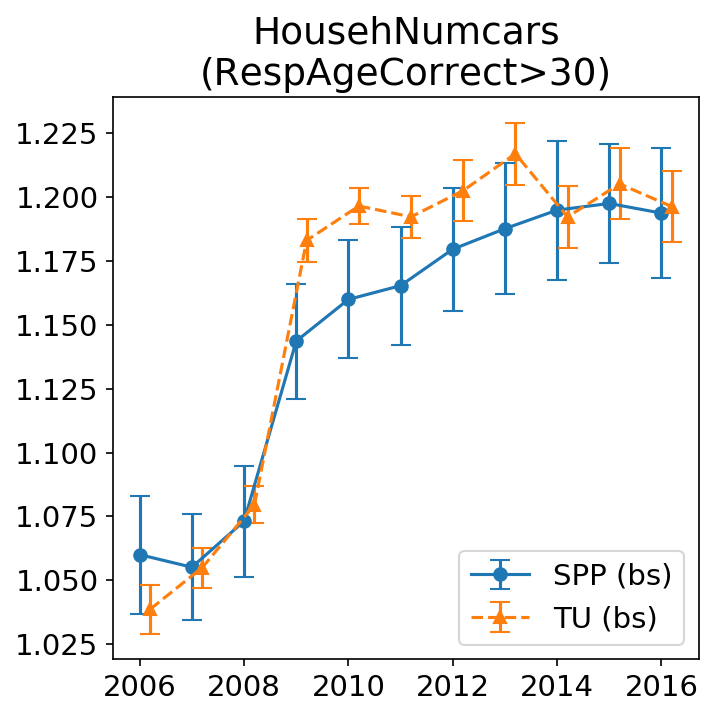}
\includegraphics[height=4.0cm]{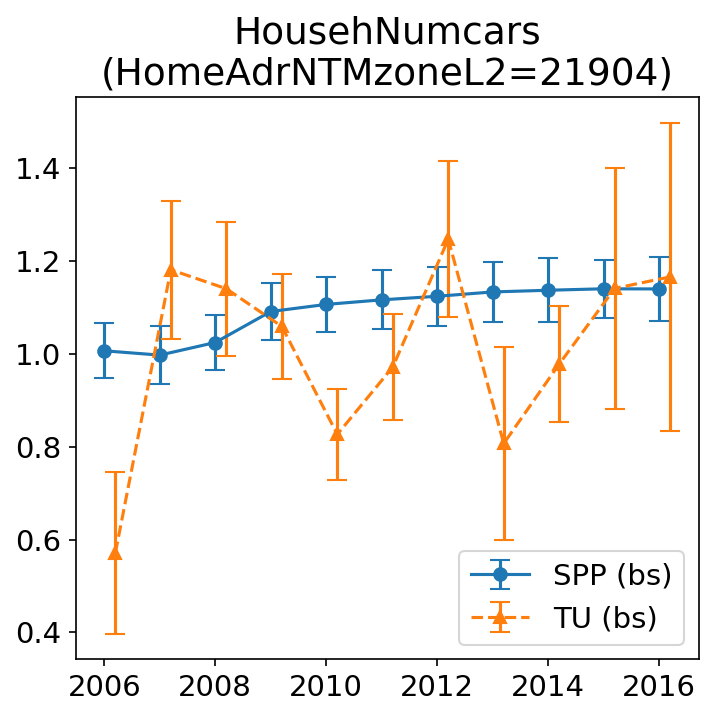}
\includegraphics[height=4.0cm]{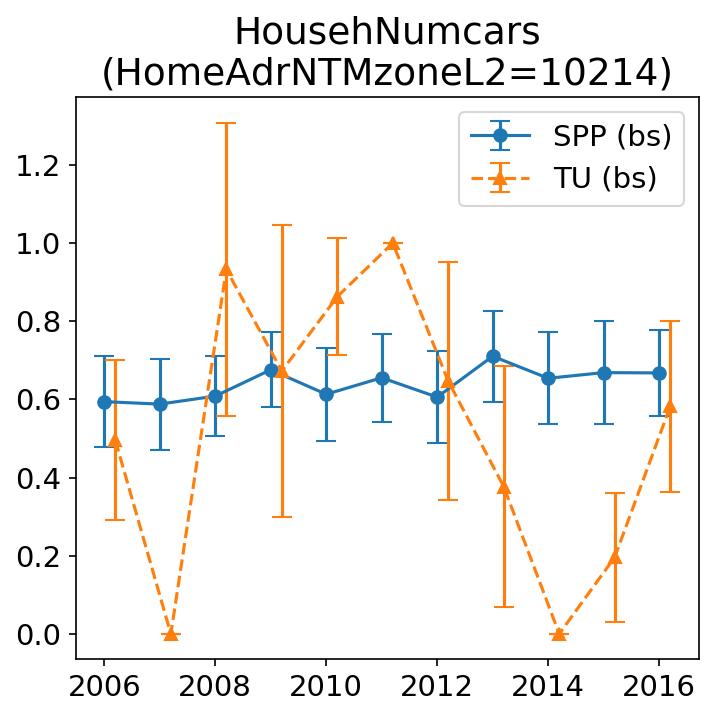}\\
\caption{Bootstrapped distributions (means and standard deviations) of the mean for two numerical attributes for various population cohorts.}
\label{fig:pref_cond_popreal_year_cond_bs}
\end{figure}

To further support these findings, we provide a statistical bootstrapping analysis. For the TU data, we sample with replacement the whole data set 100 times and calculate bootstrapped statistics for the travel preferences. For the SPP, we sample with replacement the whole data set 100 times and fit a CVAE to each of the sampled data sets. We then generate 100 travel preference samples from each of the 100 CVAEs and calculate bootstrapped statistics. According to the results shown in Fig.~\ref{fig:pref_cond_popreal_year_cond_bs}, two important observations can be made. Firstly, that the SPP provides stable estimates independently of the number of observations in a given cohort. The model noise resulting from learning different models on different held-in data is unsystematic and resembles white noise. Secondly, the bootstrapped standard errors of the TU data become highly volatile when the number of observations for a given combination of attributes decreases. This leads to poor estimates for the detailed cohorts, while, at the same time, SPP provides stable and reliable estimates.

\begin{figure}[t!]
\centering
\includegraphics[height=3.2cm]{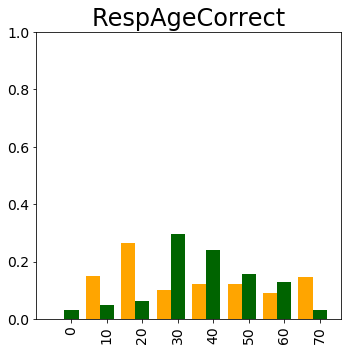}
\includegraphics[height=3.2cm]{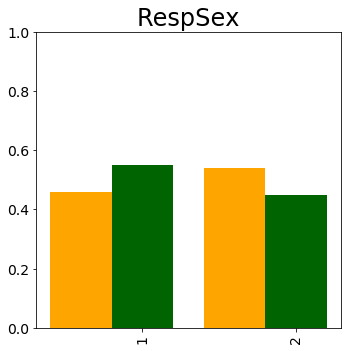}
\includegraphics[height=3.2cm]{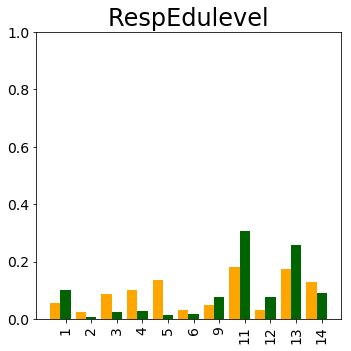}
\includegraphics[height=3.2cm]{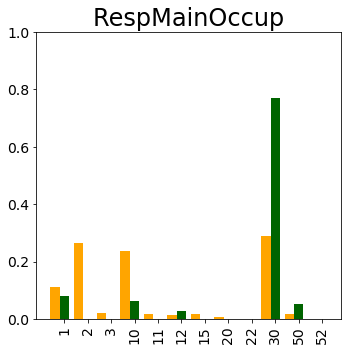}
\includegraphics[height=3.2cm]{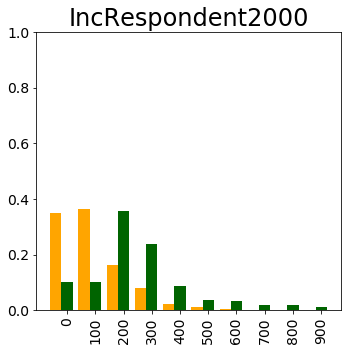}\\
\includegraphics[height=3.2cm]{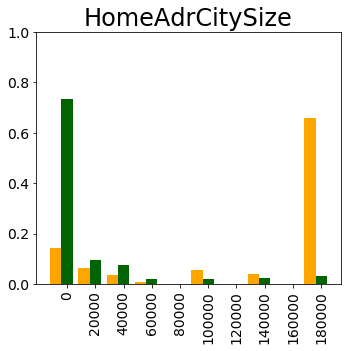}
\includegraphics[height=3.2cm]{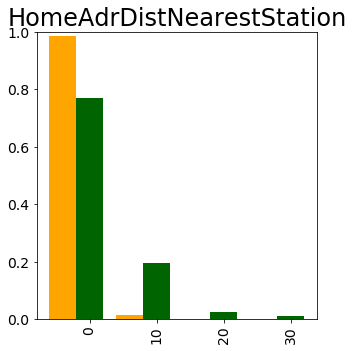}
\includegraphics[height=3.2cm]{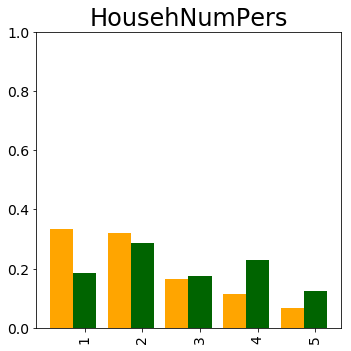}
\includegraphics[height=3.2cm]{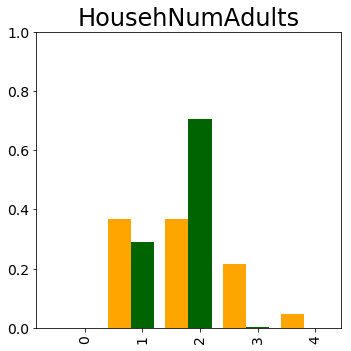}
\includegraphics[height=3.2cm]{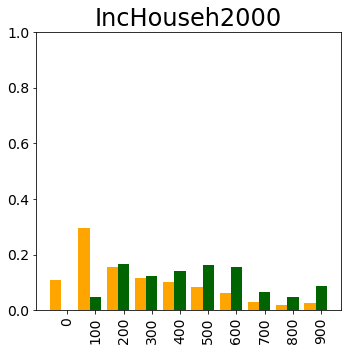}\\
\includegraphics[height=3.2cm]{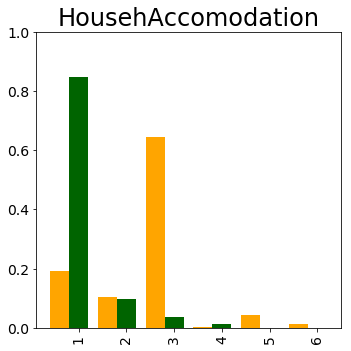}
\includegraphics[height=3.2cm]{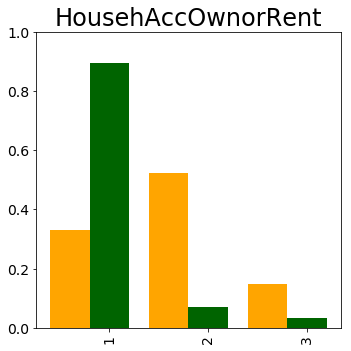}
\includegraphics[height=3.2cm]{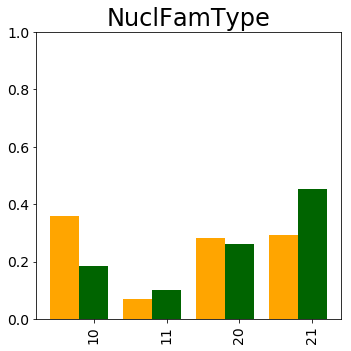}
\includegraphics[height=3.2cm]{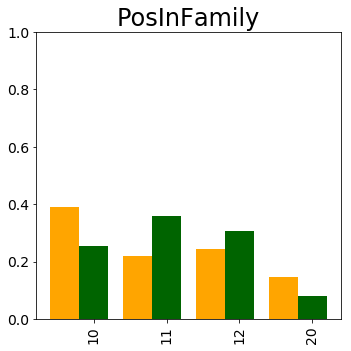}
\caption{Marginal distributions of socio-economic attributes of the fast (orange) and slow (green) movers in the travel preference space defined by the top / bottom deciles of the SRMSE distance between preference distributions of 2006 versus 2016. Young females in large cities changed their preferences most while older males in rural areas changed least. See a more detailed description in Tables~\ref{tab:fast-slow:explanation_num} and \ref{tab:fast-slow:explanation_cat}.}
\label{fig:pref_marg_fastslow}
\end{figure}

The use of the SPP also makes it possible to address other research questions, for example, to compare socio-economic profiles of people based on how much their travel preferences change during the observed period. One interesting analysis arises if people are classified according to the speed at which their preferences change. In other words, a classification of 'slow' and 'fast' movers with respect to preference changes from 2006 to 2016. We do so by calculating the SRMSE distance (Eq.~\ref{eq:srmse}) between $P_{t=2006,i}(V)$ and $P_{t=2016,i}(V)$ for all individuals. Then we range the individuals by this distance and define slow movers as those belonging to the first decile of the distance distribution whereas fast movers are those belonging to the last decile. The marginal distributions of different socio-economics attributes for these two groups are shown in Fig.~\ref{fig:pref_marg_fastslow} and described in Tables~\ref{tab:fast-slow:explanation_num} and \ref{tab:fast-slow:explanation_cat}. According to the differences between these distributions, a few observations can be made. Firstly, the prototypical fast movers, defined by the distribution mode, are young single female adults living in cities, whereas slow movers are mainly represented by middle-aged men with high incomes who live in non-single households outside the cities and in privately owned detached houses. It is also interesting to observe that elderly people over 70 years old change their preferences faster than middle-aged people. This can be related to the socio-economic developments (e.g., higher income) and accessibility improvements (e.g., better public transport). Secondly, personal income and household size are positively correlated with the probability of being a slow mover. To some extent, it is natural to expect people with high income to be less affected by societal and technological changes. Almost all students are fast movers, while employed persons are much more reluctant to change their preferences fast. Finally, almost all slow movers live in rural areas while fast movers are city dwellers. This is an expected observation given the highly dynamic changes in modern urban areas. Although no previous analysis has been carried out with the same degree of detail, it is the authors' impression that the above results corresponds well with other findings from social science.

\begin{table}[t!]
\footnotesize
\centering
\begin{tabular}{lllll}
\hline
Attribute & Values & Description & $f_\mathrm{fast}$ & $f_\mathrm{slow}$ \\
\hline
RespAgeCorrect & $[0, 10)$ & Years & 0.001 & 0.031 \\
 & $[10, 20)$ & & \bf{0.151} & 0.049 \\
 & $[20, 30)$ & & \bf{\underline{0.263}} & 0.064 \\
 & $[30, 40)$ & & 0.101 & \bf{\underline{0.295}} \\
 & $[40, 50)$ & & 0.121 & \bf{0.239} \\
 & $[50, 60)$ & & 0.122 & \it{0.158} \\
 & $[60, 70)$ & & 0.091 & 0.129 \\
 & $\geq 70$ & & \it{0.146} & 0.031 \\
\hline
IncRespondent2000 & $[0,100)$ & 1000 DKK & \bf{0.350} & 0.099 \\
 & $[100, 200)$ & & \bf{\underline{0.363}} & \it{0.101} \\
 & $[200, 300)$ & & \it{0.164} & \bf{\underline{0.356}} \\
 & $[300, 400)$ & & 0.078 & \bf{0.239} \\
 & $[400, 500)$ & & 0.022 & 0.085 \\
 & $[500, 600)$ & & 0.011 & 0.035 \\
 & $[600, 700)$ & & 0.004 & 0.032 \\
 & $[700, 800)$ & & 0.001 & 0.018 \\
 & $[800, 900)$ & & 0.001 & 0.018 \\
 & $\geq 900$ & & 0.001 & 0.011 \\
\hline
HomeAdrCitySize & $[0, 20)$ & 1000 people & \bf{0.141} & \bf{\underline{0.733}} \\
 & $[20, 40)$ &  & \it{0.062} & \bf{0.095} \\
 & $[40, 60)$ &  & 0.035 & \it{0.075} \\
 & $[60, 80)$ &  & 0.007 & 0.019 \\
 & $[80, 100)$ &  & 0 & 0 \\
 & $[100, 120)$ &  & 0.054 & 0.019 \\
 & $[120, 140)$ &  & 0 & 0 \\
 & $[140, 160)$ &  & 0.041 & 0.024 \\
 & $[160, 180)$ &  & 0 & 0 \\
 & $\geq 180$ &  & \bf{\underline{0.657}} & 0.031 \\
\hline
HomeAdrDistNearestStation & $[0, 10)$ & km & \bf{\underline{0.984}} & \bf{\underline{0.770}} \\
 & $[10, 20)$ & & \bf{0.014} & \bf{0.194} \\
 & $[20, 30)$ & & 0 & \it{0.025} \\
 & $\geq 30$ & & \it{0.001} & 0.009 \\
\hline
HousehNumPers & 1 & people & \bf{\underline{0.333}} & \it{0.185} \\
 & 2 & & \bf{0.319} & \bf{\underline{0.285}} \\
 & 3 & & \it{0.164} & 0.174 \\
 & 4 & & 0.115 & \bf{0.229} \\
 & $\geq 5$ & & 0.067 & 0.125 \\
\hline
HousehNumAdults & 0 & people & 0 & 0 \\
 & 1 & & \bf{0.368} & \bf{0.289} \\
 & 2 & & \bf{\underline{0.369}} & \bf{\underline{0.706}} \\
 & 3 & & \it{0.216} & \it{0.004} \\
 & $\geq 4$ & & 0.045 & 0 \\
\hline
IncHouseh2000 & $[0,100)$ & 1000 DKK & 0.109 & 0.001 \\
 & $[100, 200)$ & & \bf{\underline{0.295}} & 0.047 \\
 & $[200, 300)$ & & \bf{0.155} & \bf{\underline{0.165}} \\
 & $[300, 400)$ & & \it{0.116} & 0.122 \\
 & $[400, 500)$ & & 0.101 & 0.139 \\
 & $[500, 600)$ & & 0.082 & \bf{0.164} \\
 & $[600, 700)$ & & 0.061 & \it{0.156} \\
 & $[700, 800)$ & & 0.031 & 0.065 \\
 & $[800, 900)$ & & 0.018 & 0.048 \\
 & $\geq 900$ & & 0.027 & 0.088 \\
\hline
\end{tabular}
\caption{Distributions of the discretized numerical attributes of the fast and slow movers in the travel preference space. Modes are highlighted with bold font and underlined. Second and third most frequent values are highlighted with bold and italic fonts respectively.}
\label{tab:fast-slow:explanation_num}
\end{table}

\begin{table}[t!]
\footnotesize
\centering
\begin{tabular}{lllll}
\hline
Attribute & Values & Description & $f_\mathrm{fast}$ & $f_\mathrm{slow}$ \\
\hline
RespSex & 1 & Male & \bf{0.459} & \bf{\underline{0.550}} \\
 & 2 & Female & \bf{\underline{0.540}} & \bf{0.449} \\
\hline
RespEdulevel & 1 & 1st-7th form & 0.057 & \it{0.101} \\
 & 2 & 8th form & 0.022 & 0.007 \\
 & 3 & 9th form & 0.087 & 0.025 \\
 & 4 & 10th form & 0.099 & 0.027 \\
 & 5 & Upper secondary certificate, higher preparatory certificate & \it{0.135} & 0.014 \\
 & 6 & Higher commercial certificate, higher technical certificate, & 0.032 & 0.018 \\
 &   & business college &  &  \\
 & 9 & Other schooling & 0.049 & 0.075 \\
 & 11 & Vocational (certificate of apprenticeship, etc.) & \bf{\underline{0.179}} & \bf{\underline{0.305}} \\
 & 12 & Short-term further education (1.5 - 2 years) & 0.031 & 0.078 \\
 & 13 & Medium-term further education (2 - 5 years) & \bf{0.174} & \bf{0.256} \\
 & 14 & Long-term further education (minimum 5 years) & 0.129 & 0.089 \\
\hline
RespMainOccup & 1 & Pupil & 0.111 & \bf{0.081} \\
 & 2 & Student & \bf{0.265} & 0.001 \\
 & 3 & Apprentice, trainee & 0.021 & 0.001 \\
 & 10 & Retired person, state pension, early retirement pension & \it{0.235} & \it{0.062} \\
 & 11 & Unemployed & 0.018 & 0.001 \\
 & 12 & Receiver of pre-retirement pay & 0.014 & 0.028 \\
 & 15 & Social assistance, rehabilitation, long-term ill & 0.018 & 0 \\
 & 20 & Full-time housewife/-husband, otherwise out of work & 0.007 & 0.001 \\
 & 22 & National serviceman & 0 & 0 \\
 & 30 & Employee & \bf{\underline{0.289}} & \bf{\underline{0.770}} \\
 & 50 & Self-employed & 0.018 & 0.051 \\
 & 52 & Assisting spouse (of self-employed person) & 0 & 0 \\
\hline
HousehAccomodation & 1 & Detached single-family house & \bf{0.191} & \bf{\underline{0.848}} \\
 & 2 & Terraced house, linked house & \it{0.104} & \bf{0.098} \\
 & 3 & Block of flats & \bf{\underline{0.643}} & \it{0.038} \\
 & 4 & Farm & 0.002 & 0.012 \\
 & 5 & Student residence & 0.044 & 0 \\
 & 6 & Other & 0.014 & 0.001 \\
\hline
HousehAccOwnorRent & 1 & Owner-occupied dwelling & \bf{0.329} & \bf{\underline{0.895}} \\
 & 2 & Rent & \bf{\underline{0.522}} & \bf{0.071} \\
 & 3 & Cooperative & \it{0.148} & \it{0.032} \\
\hline
NuclFamType & 10 & Single & \bf{\underline{0.358}} & \it{0.185} \\
 & 11 & Single with child/children & 0.068 & 0.099 \\
 & 20 & Couple & \it{0.281} & \bf{0.262} \\
 & 21 & Couple with child/children & \bf{0.292} & \bf{\underline{0.452}} \\
\hline
PosInFamily & 10 & Single & \bf{\underline{0.390}} & \it{0.253} \\
 & 11 & Older in couple & \it{0.221} & \bf{\underline{0.358}} \\
 & 12 & Younger in couple & \bf{0.242} & \bf{0.306} \\
 & 20 & Child in nuclear family (under 25 years of age) & 0.145 & 0.081 \\
\hline
\end{tabular}
\caption{Distributions of the categorical attributes of the fast and slow movers in the travel preference space. Modes are highlighted with bold font and underlined. Second and third most frequent values are highlighted with bold and italic fonts, respectively.}
\label{tab:fast-slow:explanation_cat}
\end{table}

%-------------------------------------------------------------------------

\section{Summary and conclusion}
\label{sec:conclusion}

In the paper, a new methodology for constructing heterogeneous pseudo panels from repeated cross-sectional data is presented. The approach is based on estimating the joint distribution of travel preferences represented by a number of transport-related attributes but conditional on a set of socio-economic variables and other external variables related to e.g. infrastructure and external conditions. The probabilistic framework allows sampling from the travel preference distribution conditioned on heterogeneous cohorts of individuals. This in turn makes it possible to circumvent a well-known aggregation problem when constructing cohorts for pseudo-panel analysis. Technically, the estimation of the conditional distribution is based on a Conditional Variational Autoencoder (CVAE) framework, which is a generative model based on an artificial neural network. The model is able to mimic the properties of high-dimensional surveys and can support the creation of pools of individuals based on random sampling from the model. Individuals within the sampled pool will be comparable to the individuals in the original survey in a statistical sense in that all of the intrinsic correlation properties between attributes are preserved. However, the sampled individuals will not be strict copies but imputed and learned from the underlying distribution. There are many potential applications of detailed pseudo panels within transport, medicine, bio-statistics, finance and economics. Moreover, the approach may also be relevant as a means to model population synthesis and to address data privacy issues.

As a study case, we analysed repeated cross-sectional data from the Danish National Travel Survey (TU). We show that the CVAE indeed captures statistical properties of the data and is a valuable tool for tracking detailed dynamic preferences. A thorough validation of the model is provided including early stop assessment, goodness-of-fit performance and bootstrapping of model performance in order to assess dependencies with respect to the 'held-in' data. It is shown that the choice of specific 'held-in' data has little influence on model performance and that the bootstrapped variance is generally low and unsystematic. Moreover, the revealed dynamic travel preferences for the SPP are shown to conform well to that of the TU participants, with the exception that the trends revealed on the basis of the TU are more volatile due to the presence of sampling noise and fluctuations in the socio-economic variables that arise from structural changes (e.g., the financial crisis or economic growth). The estimation of the preference probability distribution jointly for all years and geographic locations allows for different ways of data utilization because the data can be specifically aligned depending on the design and purpose of the study. Also, it makes it possible to explore very sparse data-spaces through the indirect imputation of data-points facilitated by the CVAE model.

To illustrate the potential of the SPP approach, an analysis of 'fast' and 'slow' movers in the travel preference space is provided. The two groups are defined as those groups of people who changed their preferences the most and the least, respectively, over the period of 11 years from 2006 to 2016. Through the heterogeneous description of individuals it is possible to reveal a very detailed socio-economic classification of fast and slow movers. It is found that the prototypical fast movers are young single females who live in cities, whereas slow movers are primarily represented by middle-aged men with high incomes who live in non-single households outside the cities in privately owned detached houses. We also found an indication that elderly people over the age of 70 are changing their preferences faster than middle-aged people and confirmed that fast movers are predominantly city dwellers. Such investigations are, to our best knowledge, of general interest to the transport research community.

\subsection{Future research}
\label{sec:conclusion:future}

Although the presented methodology has a great potential, there are several directions for future research.

An important topic is that of model validation. When investigating high-dimensional distributions, which are not in a human-readable format, model validation becomes an extremely challenging task. This paper provides some extensions to common practise by looking at held-in data robustness through bootstrapping. However, a rigorous model validation paradigm for evaluating deep generative models is to be developed. 

From an application perspective, the presented case study analysis can be further extended. For instance, it would be interesting to investigate preference dynamics in more detail using factor models or to include a more detailed analysis of socio-economic attributes of fast/slow movers using clustering. Exploring the latent space of CVAE models can also reveal various useful properties of the data. As similar data points tend to be close to each other, clusters in the latent space may also provide useful insights.% Various dimensions of the latent space can have meaning. For example, for images of hand written symbols, it can be their thickness, oblique or size. However, as the data is nonlinearly transformed, the latent variables should be explored manually to assign any meaning to them. From this point of view, it would be interesting to explore the latent space for the modelled preference and socio-econiomic attributes.

%-------------------------------------------------------------------------

\section*{Acknowledgements}

The research leading to these results has received funding from the European Union's Horizon 2020 research and innovation programme under the Marie Sklodowska-Curie grant agreement no.~713683 (COFUNDfellowsDTU). The authors also thank Mogens Fosgerau for useful discussions.

%-------------------------------------------------------------------------



%% References
%%   \citep{key}          ==>>  [#]


\bibliographystyle{elsarticle-harv}
\bibliography{spp.bib}






% \end{thebibliography}

\newpage

% \begin{figure}[ht]
% \centering
% \includegraphics[height=7.0cm]{{"img/pp"}.png}
% \caption{Construction of preference pseudo-panel.}
% \label{fig:pp}
% \end{figure}

% \begin{figure}[ht]
% \centering
% \includegraphics[height=6.0cm]{img/cvae_train}
% \hspace{1.5cm}
% \includegraphics[trim={2.0cm 0.0cm 2.0cm 0.0cm},clip,height=6.0cm]{img/cvae_sample}
% %\includegraphics[height=4.0cm]{img/cvae_sample}
% \caption{Conditional Variational Autoencoder.}
% \label{fig:cvae}
% \end{figure}

\end{document}